\documentclass[10pt,twocolumn,letterpaper]{article}

\usepackage{iccv}
\usepackage{times}
\usepackage{epsfig}
\usepackage{graphicx}
\usepackage{amsmath}
\usepackage{amssymb}

\usepackage{bm}
\usepackage{color}
\usepackage{subfig}

\usepackage{amssymb}
\usepackage{bm}
\usepackage{scalerel}

\usepackage{array,multirow,graphicx}
\usepackage{float}
\usepackage{subfig}
\usepackage{array}
\usepackage{makecell}

\newcolumntype{P}[1]{>{\centering\arraybackslash}p{#1}}

\makeatletter
\renewcommand{\maketag@@@}[1]{\hbox{\m@th\normalsize\normalfont#1}}%
\makeatother

\usepackage[pagebackref=true,breaklinks=true,letterpaper=true,colorlinks,bookmarks=false]{hyperref}

\iccvfinalcopy 

\def\iccvPaperID{6264} 

\title{Can generalised relative pose estimation solve sparse 3D registration?}

\author{
    Siddhant Ranade$^{1*}$ \quad Xin Yu$^{1}\thanks{$^{*}$indicate equal contributions.}$ \quad Shantnu Kakkar$^{1}$ \quad Pedro Miraldo$^{2}$ \quad Srikumar Ramalingam$^{1}$
 \\$^1$ University of Utah \quad $^2$ KTH Royal Institute of Technology
 \\{\tt\small \{xiny,sidra,srikumar\}@cs.utah.com, s.kakkar@utah.edu, miraldo@kth.se
}
}

\begin{document}

\maketitle

\begin{abstract}
Popular 3D scan registration projects, such as Stanford digital Michelangelo or KinectFusion, exploit the high resolution sensor data for scan alignment. It is particularly challenging to solve the registration of sparse 3D scans in the absence of RGB components. In this case, we can not establish point correspondences since the same 3D point can not be captured in two successive scans. In contrast to correspondence based methods, we take a different viewpoint and formulate the sparse 3D registration problem based on the constraints from the intersection of line segments from adjacent scans. We obtain the line segments by modeling every horizontal and vertical scan-line as piece-wise linear segments. We propose a new alternating projection algorithm for solving the scan alignment problem using line intersection constraints. We develop two new minimal solvers for scan alignment in the presence of plane correspondences: 1) 3 line intersections and 1 plane correspondence, and 2) 1 line intersection and 2 plane correspondences. We outperform other competing methods on Kinect and LiDAR datasets.
\end{abstract}

\section{Introduction}
The last few years have witnessed the rise of inexpensive 3D sensors for both indoor (e.g., Microsoft Kinect) and outdoor scenes (e.g., Velodyne LIDAR’s VLP-16 LITE and ZED stereo camera). One of the key problems in working with 3D sensors is the alignment of scans from different viewpoints and orientations. Most algorithms rely on obtaining explicit point correspondences using associated RGB components, or dense point-cloud with good initialization for jump-starting the iterative closest point (ICP) algorithm. In this work we study the following problem: How do we align highly sparse 3D scans without RGB information? Note that it is not possible to obtain point correspondences from sparse 3D data since we don't observe the same 3D point in two consecutive scans, and we may not always have sufficient plane correspondences. Thus, it is not possible to directly apply ICP methods. In order to handle this challenge, we take a different viewpoint in solving the registration problem.

\begin{figure}[t]
    \begin{center}
        \includegraphics[width=\linewidth]{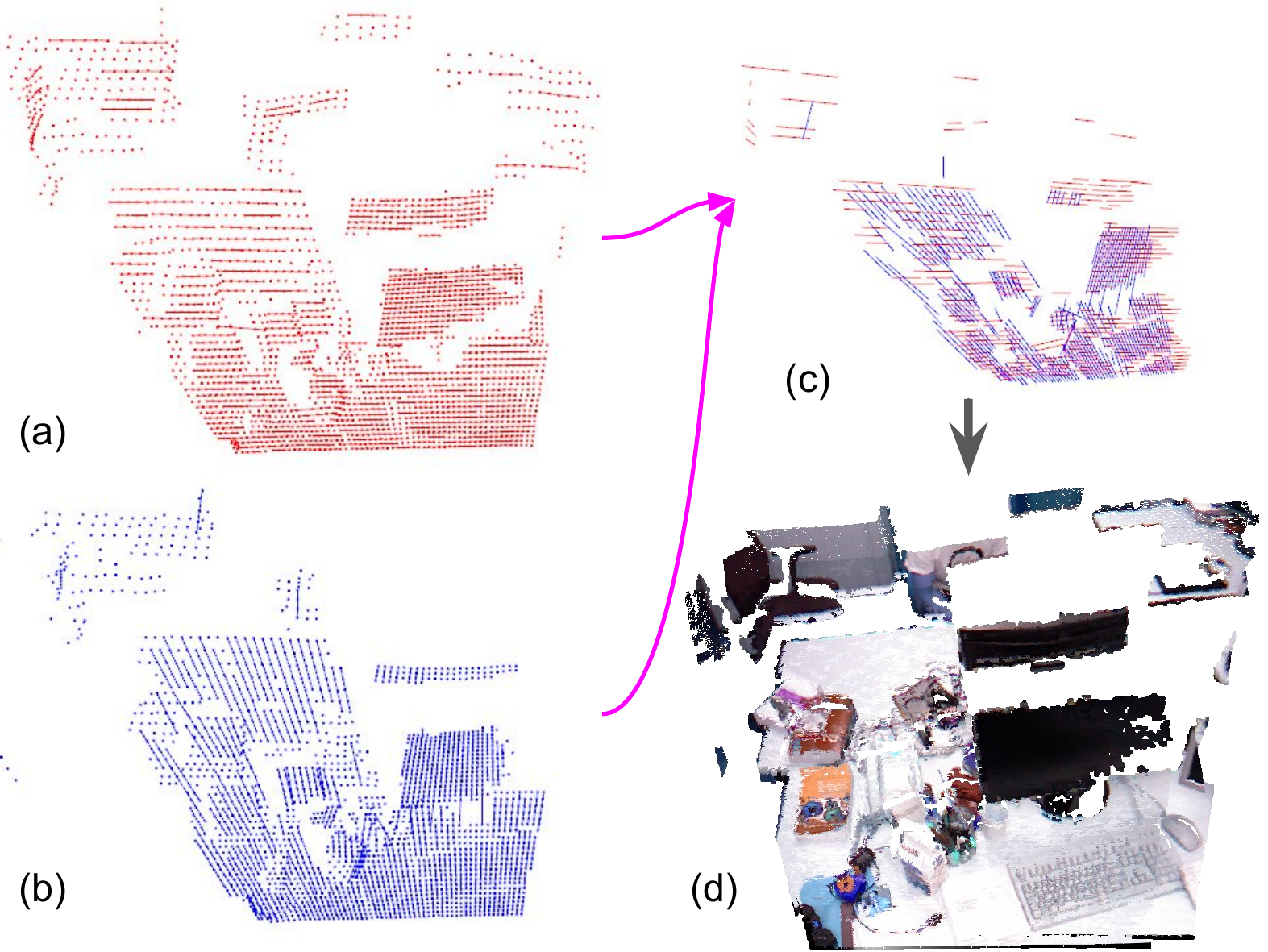}
    \end{center}
    \caption{We show 100x down-sampled Kinect depth data (64 x 48) without RGB for two scans.We fit line segments to rows in the first frame, and columns in the second frame as shown in (a) and (b), respectively. We identify potential line-to-line intersection constraints between the frames. We use alternating projection to solve the line-to-line intersection constraints to obtain an accurate scan alignment or the relative pose as shown in (c). (d) shows the registered point-cloud in original resolution with color using the computed relative pose. The average error is given by 0.5212 degrees for rotations, and 5 mm for the translation. The standard 3-point minimal solver with RANSAC achieves an error of 0.740 for rotation, and 24 mm for translation on the original scans along with using RGB components. We outperform standards methods by a large margin, despite using 100X less points and without relying on RGB components.} 
    \label{fig:intro_figure}
\end{figure}

We show the basic idea behind our registration algorithm in Fig.~\ref{fig:intro_figure}. We consider 100X down-sampled Kinect data without RGB components. We show that the proposed algorithm can outperform competing Kinect registration algorithms that use full resolution and RGB components. In the geometric vision community, there have been a wide variety of registration algorithms that are either minimal or non-minimal. Most of these registration problems can be solved once we obtain the formulation and identification of the underlying constraints. The real exciting part is the identification of the equivalence between seemingly different computer vision problems. Now consider the following question:

\begin{itemize}
    \item Can the solution to generalized relative pose estimation~\cite{pless03,Stewenius2005,sturm05,Li2008} help to solve the sparse scan alignment problem?
\end{itemize}

\begin{figure}[!htbp]
    \begin{center}
        \psfig{figure=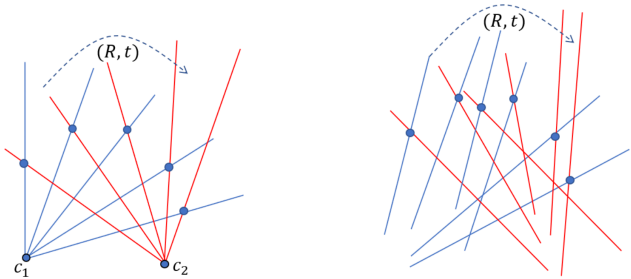,width=1.0\columnwidth}
    \end{center}
    \caption{On the left, we show the intersection of five pairs of projection rays from a perspective camera for relative pose estimation. On the right, we show the intersection of six pairs of unconstrained projection rays from a generalized cameras for the relative pose estimation.} 
    \label{fig.relative_pose_intro}
\end{figure}

\noindent 
\textbf{Transforming scan alignment to generalized relative pose estimation problem:} We show that the registration of sparse point-clouds can be mapped to the relative pose estimation from cameras, more specifically, generalized cameras~\cite{pless03,Stewenius2005,sturm05,Li2008}. The problem of relative pose estimation involves finding rotation and translation between two cameras such that the projection rays associated with corresponding 2D points intersect with each other (See Fig.~\ref{fig.relative_pose_intro}). In the case of perspective cameras, we can find the motion up to a scale, and the associated projection rays are central~\cite{nister03}. In the case of generalized cameras~\cite{grossberg01}, the projection rays are unconstrained and we are looking at the alignment of two sets of line segments such that the corresponding line segments intersect with each other. 

It may appear that we have solved the alignment problem since generalized relative motion estimation is a well studied problem. While minimal solvers typically produce robust solutions in general, the 6-point minimal solver for generalized relative pose \cite{stewenius05} is plagued by too many degeneracy problems. For example, if multiple unconstrained projection rays pass through a single line in space, then it is difficult to extract the unique pose. Furthermore, we obtain 64 solutions and we need to find the correct solution using other correspondences. In this work, we solve this problem using alternating projection method with 7 pairs to avoid degeneracy issues. 

Overall, this paper investigates the possibility of using line intersection constraints, rather than traditional point or line correspondences, for solving the registration problem. Nevertheless, we will still exploit planar correspondences in scenarios where they are available. We summarize the contributions of this paper:

\begin{enumerate}
\item We propose an alternating projection algorithm for the problem of scan alignment by showing that it is equivalent to the generalized relative pose estimation, i.e., pose estimation by satisfying intersection constraints among 6 or more pairs of corresponding line segments.
\item We propose two new minimal solvers for solving the scan alignment in the presence of plane correspondences:
(a) One plane correspondence and 3 line intersection constraints (3L1P).
(b) Two plane correspondences and 1 line intersection constraint (1L2P).
\item We outperform other registration algorithms for Kinect, despite using highly down-sampled point-cloud and not relying on the color information.
\item We outperform competing LiDAR registration algorithms such as LOAM~\cite{Zhang14} that performs distortion compensation from moving platforms, and minimizes point-to-line and point-to-plane distances in a Levenberg Marquardt framework. 
\end{enumerate}

\section{Related Work}

3D scan alignment is a classical problem in the computer vision and robotics communities, and there is a rich body of literature on this topic. Iterative Closest Point (ICP) \cite{arun87,horn87,umeyama91,besl92,penney01} is the most applied scan alignment algorithm and it works well for dense point-clouds with good initialization, and several variants including planar ones have been proposed \cite{ROB-035}. 

\vspace{.25cm}
\noindent{\bf 3D SLAM:} Other popular scan alignment methods in the absence of good initialization include minimal 3-point solvers in a RANSAC framework. Most common approaches to remove outliers from the data are based on the use of RANSAC~\cite{fischler81} plus some three 3D point correspondences solver such as the Procrustes's solvers \cite{schonemann66}. In addition to points, several registration algorithms have utilized other features on beam-based environment modeling~\cite{endres14}, 3D planes~\cite{Zhou-2018-107715,ma16,bhattacharya17,liu18,Grant2018}, 3D line segments~\cite{lu15,Zhou-2018-107715}, implicit surface representation~\cite{Deschaud18}, and edges~\cite{choi13}. A detailed survey on 3D SLAM methods can be found in ~\cite{endres12,sturm12}.
 
Techniques to find globally optimal solutions combine local or probabilistic methods with graph optimization \cite{THEILER2015126} and branch-and-bound \cite{Campbell16,li17}. A closely related paper to our work is LOAM~\cite{Zhang14,Zhang17}, which is a top-ranking LiDAR alignment algorithm. We differ with LOAM in the following. First, LOAM uses a distortion correction step for addressing the motion of the sensor at low frame-rate. This is similar to the rolling shutter effect in cameras. Second, LOAM uses point-to-line and point-to-plane distances, while we use line intersection constraints. Third, LOAM uses the Levenberg Marquardt method, which is prone to local minima issues, while we develop alternating minimization and algebraic minimal solvers instead, and utilize the RANSAC framework. Fourth, LOAM typically (but optionally) uses IMU and other sensors, and we rely only on sparse 3D points. Finally, our method does not rely on any boundary or edge points, while LOAM explicitly identifies edge points in the registration algorithm. 

Several global methods have been proposed in the literature, such as~\cite{yang13,yang16}.
One of the problems with these methods is the high computation requirement for doing the branch and optimization. To overcome this, one line of research attempts to decompose the task of finding the relative transformation into first finding the rotation and then obtaining the translation given the optimal rotation \cite{Makadia06_cvpr, Straub17_cvpr}. Further, \cite{Liu18_eccv} proposes Rotation Invariant Features (RIF) to ease the task of decoupling rotation solution from the translation. However, none of the methods have been tested on LiDAR data.

\vspace{.25cm}
\noindent
{\bf Alternating projection:}
In this work, we develop a near-minimal solver using alternating projection (AP) algorithm. Alternating minimization and projection algorithms have been used in clustering, non-negative matrix factorization, dictionary learning, and many other problems. In 3D computer vision, alternating minimization methods have been used for many problems including shape from templates ~\cite{Zhou2015a,Zhou2015b}, and multi-view point registration~\cite{yan15}.

\vspace{.25cm}
\noindent
{\bf Minimal Solvers:}
Our approach for the registration of 3D scans is tightly connected with minimal solvers for relative pose estimation using points, lines, and planes. In particular, we have relative pose estimation algorithms for calibrated perspective cameras~\cite{nister03,li06b}, with known relative rotation angle~\cite{li13}, with known directions~\cite{fraundorfer10,saurer15}, with unknown focal lengths~\cite{stewenius05b,li06}, solutions invariant to translation~\cite{kneip12}, and generalized relative pose~\cite{stewenius05c,ventura15}. Recently, a hybrid minimal solver considers both relative and absolute poses as shown in  \cite{camposeco18}. In the case of absolute poses, we have for perspective ~\cite{kneip11,ke17,wang18,persson18}, and multi-perspective systems~\cite{ventura14,camposeco16}.

\vspace{.25cm}
\noindent
{\bf Deep scan alignment:}
 Deep neural networks have been used to extract local 3D geometric structures in \cite{elbaz17,khoury17}. There have been a few recent algorithms for LiDAR registration~\cite{Ding2019,Elbaz2017} using deep neural networks, but they are mostly applicable on dense point-clouds. Floor plan reconstruction using deep networks has been shown in~\cite{liu18}.

\section{Problem Statement and Roadmap}
Given two scans, each with a sparse set of 3D points without RGB information, we are interested in computing the motion ($\mathbf{R},\mathbf{t})$ that can align two scans into the same reference frame. The main steps in our registration algorithm (See Fig.~\ref{fig:intro_figure}) is as following:

\begin{enumerate}
    \item Given a sparse set of 3D points in a regular grid (with some points missing due to sensor noise), we fit line segments to points from individual columns and rows as shown in Fig.~\ref{fig:intro_figure}. In the first frame we extract line segments on the individual columns. In the second frame we extract line segments from the individual rows.
    \item We compute potential line intersection constraints by observing the distance between line segments in the first and second scans. It is important to observe that we rely on line intersection constraints and not on line correspondences. Line intersections are many-to-many, while line correspondences are just one-to-one. This makes the line  intersection constraints easier to obtain than line correspondences.
    \item We develop a new alternating projection algorithm to register pairs of line segments from two different scans. We use this solver in a RANSAC framework to compute the pose between two adjacent scans.
    \item We also show that we can have a minimal solver in closed form when we have 1 or 2 plane correspondences, in addition to line intersection constraints. We are aware that three plane correspondences are sufficient to generate registration, but we can not always rely on the availability of 3 or more good plane correspondences from sparse noisy point-clouds. Furthermore, we need planes in many orientations to avoid degeneracy issues, and this can not always be guaranteed. 
\end{enumerate}

\section{Alternating Projection}

This is a simple constraint satisfaction algorithm for computing a point in the intersection of some sets, using a sequence of projections on to the sets. The projection of a point $x$ (in $D$ dimensions in general) on to a set ${\cal S}$ is equivalent to finding a point in the set $x' \in {\cal S}$ such that $x' = \arg\min_{x'} |x-x'|$, where $|x - x'|$ is the Euclidean distance. The associated operator $P : x \rightarrow x',x' \in {\cal S}$ is referred to as the projection. In other words, the projection operator moves the variables by the smallest amount so that they satisfy a particular constraint.

 \begin{figure}[!htbp]
    \begin{center}
        \psfig{figure=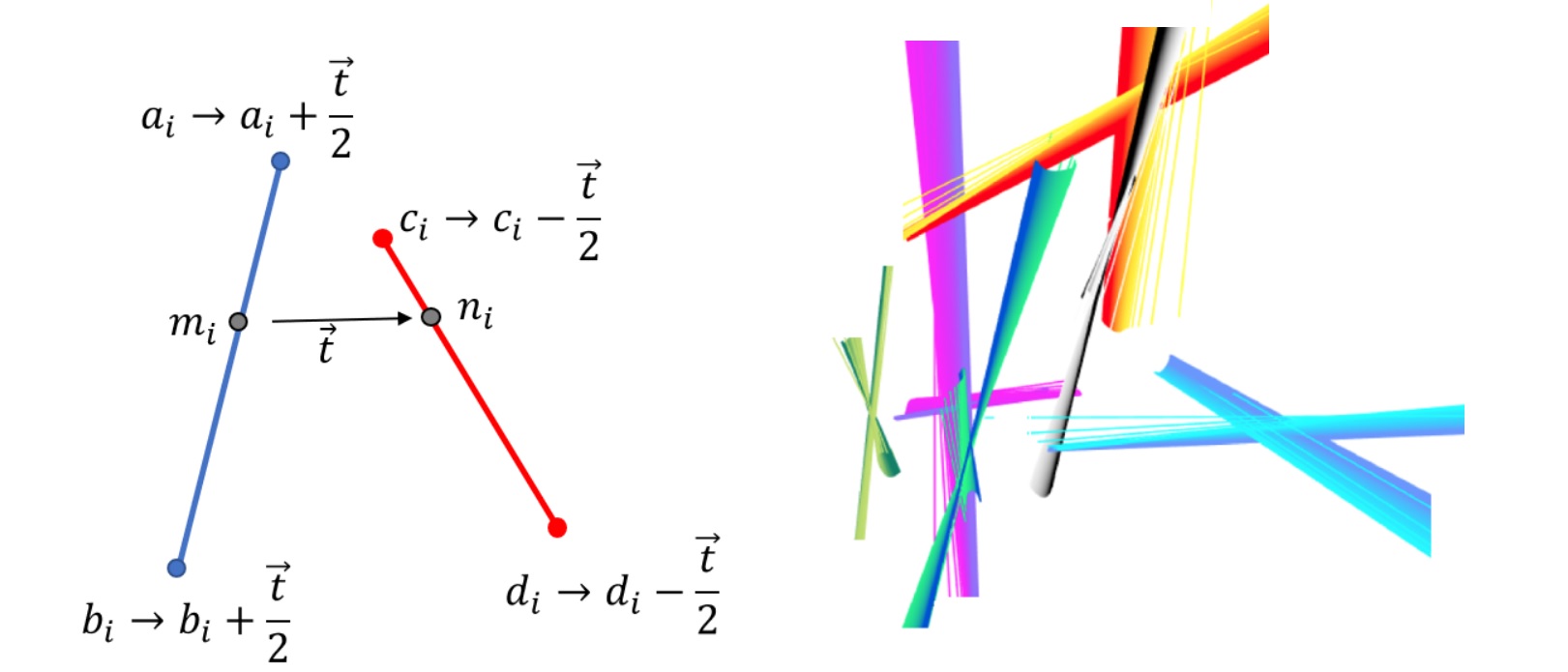,width=1.0\columnwidth}
    \end{center}
    \caption{On the left we show the projection operation for the intersection constraint. The closest points on the line segments are shown by $m_i$ and $n_i$ respectively. The end points of the line segments are moved by half the distance between these closest points to make the two line segments intersect. On the right, we show the visualization of the alternating projection algorithm for 6 pairs of line segments. The starting position of the line segments are shown in light color and the final position of the line segments are shown in dark. The line segments sweep through a surface as the endpoints move based on the projection operations for the intersection and the rigidity constraints.} 
    \label{fig:ap_vis}
\end{figure}

We formulate the problem of satisfying the intersection of corresponding line segments from two scans using two constraints: {\bf intersection}, and {\bf rigidity}. The intersection constraint moves the associated points from two corresponding line segments by a small amount so that the line segments intersect with each other. The rigidity constraint enforces the property that the collection of the end points from the line segments in each individual scan remain rigid as a whole. During the alternating projection algorithm, when we satisfy the intersection constraint, we may violate the rigidity constraint, and vice versa. 

Let us consider a set of $N$ line segments in two different frames ${\cal S}_1$ and ${\cal S}_2$, given by the pair of end points $(a_i,b_i),i=(1,...,N)$ and $(c_i,d_i),i=(1,...,N)$. The correspondences are chosen based on the inter-line distances between the line segments. The basic idea of the algorithm is to keep changing the values (projection operations to satisfy the associated constraints) of $(a_i,b_i,c_i,d_i)$ till the corresponding line segments intersect each other within a distance threshold of $\epsilon$.

Let $(a^s_i,b^s_i,c^s_i,d^s_i)$ and $(a_i,b_i,c_i,d_i)$ denote the original input values and the current updated values for the line coordinates, respectively. The algorithm is summarized below:

\begin{enumerate}
\item {\bf Intersection constraint:} Given a pair of line segments $(a_i,b_i)$ and $(c_i,d_i)$, we find the closest points on both the line segments. We define a projection operation to move the 3D coordinates of all the four points $(a_i,b_i,c_i,d_i)$ by the same distance to perfectly satisfy the intersection constraint as shown in Fig.~\ref{fig:ap_vis}. We iterate this for all the $N$ line pairs. 
\item {\bf Rigidity constraint:} Find rigid body transformations ${\cal T}_i$ and ${\cal T'}_i$ to impose the rigidity constraints for the points in both the scans as shown below:

\begin{eqnarray}
(a_i,b_i) & = & {\cal T}_i (a^s_i,b^s_i) \\
(c_i,d_i) & = & {\cal T}'_i (c^s_i,d^s_i)
\end{eqnarray}

\item Iterate steps 1 and 2 till the updates are less than a threshold $\epsilon$.
\end{enumerate}

\section{Minimal Solvers}
\label{sec:mix}

To represent 3D lines, we use {\it Pl\"ucker} coordinates \cite{pottmann01}, i.e. lines are represented by a six dimensional vector $l \in\mathbb{R}^6$, the first three elements represent the line's direction and the last three its moment vector. Planes are represented by a four-tuple $\pi \in\mathbb{R}^4$ with the first three elements corresponding to the normal of the plane, and the last element denoting the distance of the plane from the origin.

Our goal is to compute the rotation matrix $\mathbf{R}$ and translation $\mathbf{t}$. As shown in the generalized relative pose estimation problem, the constraint for the intersection of two line segments $l \in \mathbb{R}^6 $ and $m \in \mathbb{R}^6 $ in two coordinate frames with relative motion given by $(\mathbf{R},\mathbf{t})$ is as follows~\cite{pless03}:

\begin{equation}\label{eq:klien}
    m^T
    \begin{bmatrix}
    -[\mathbf{t}]_{\text{x}} \mathbf{R} & \mathbf{R} \\
    \mathbf{R} & \mathbf{0}
    \end{bmatrix}
    l .
\end{equation}

For solving both the minimal problems, we pre-process the line segments and planes to move them to a canonical coordinate frame where the relative motion between the two scans is simple, and then solve the problem. The pre-processing transformation can be reversed later to generate the actual transformation between the original scans. More details on the nature of the constraints and underlying equations will be provided in the Supplementary Materials. 
 
\subsection{One Line Intersection and Two Plane Correspondences (1L2P)} 
\label{sec:1l2p}

 \begin{figure}[ht]
    \begin{center}
        \psfig{figure=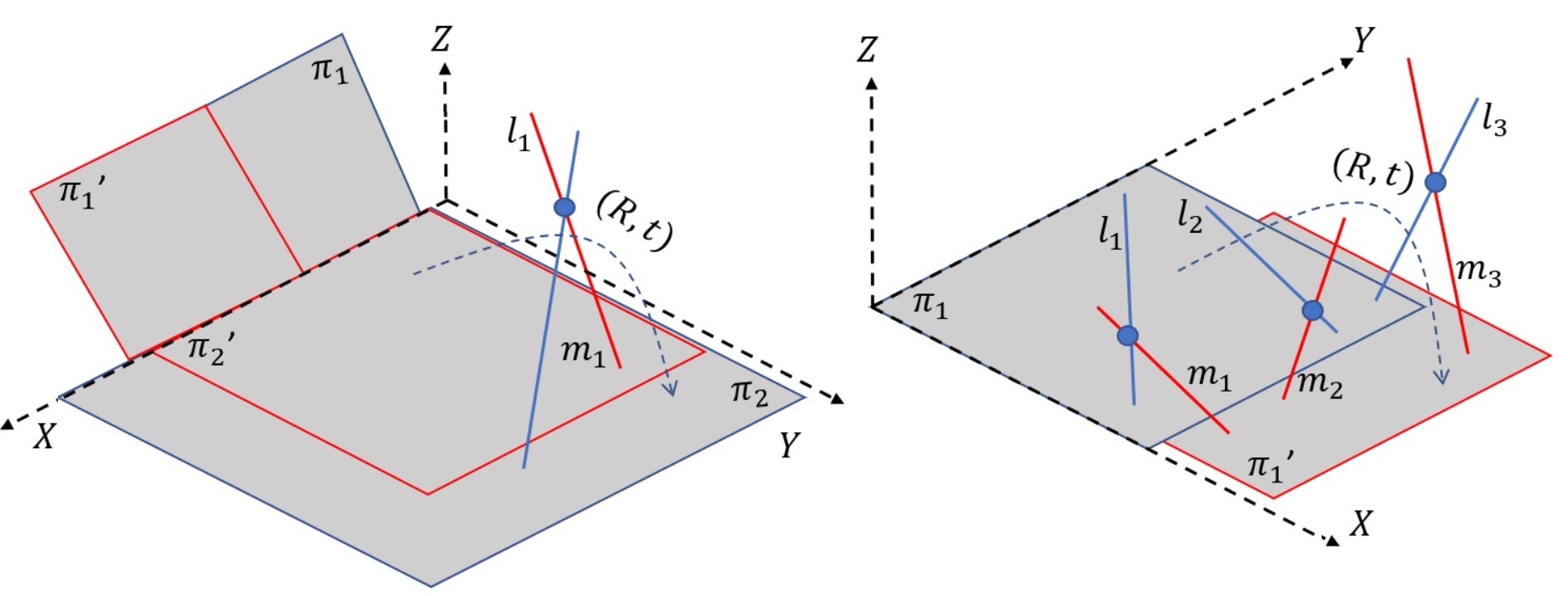,width=1.0\columnwidth}
    \end{center}
    \caption{Representation of the two minimal solvers problems solved in this paper. On the left we show the case of one line intersection and two plane matches. On the right we show the case of three line intersections and one plane match.} 
    \label{fig:minimal_solvers}
\end{figure}

As shown in figure~\ref{fig:minimal_solvers}, we are given one line intersection constraint ($l_1,m_1$) and two plane correspondences $\{(\pi_1,\pi'_1),(\pi_2,\pi'_2)\}$ from two different scans. We transform the two scans such that the following conditions are satisfied:
\begin{enumerate}
    \item The planes $\pi_2$ and $\pi'_2$ are on the $XY$ plane.
    \item The $X$ axis is along the intersection of the planes $\pi_1$ and $\pi_2$. Similarly, $X$ axis is along the intersection of the planes $\pi'_1$ and $\pi'_2$.
\end{enumerate}
After this pre-processing transformations, the relative motion $(\mathbf{R},\mathbf{t})$ between the two scans is simplified as follows:
\begin{equation}
    \label{eq:1l2p_new_Rt}
    \mathbf{R} = \mathbf{I}
    \ \ \text{and} \
    \mathbf{t} =
    \begin{bmatrix}
    t_1 & 0 & 0
    \end{bmatrix}^T.
\end{equation}
where $\mathbf{I}$ is the $3\times 3$ identity matrix. 
There is only one unknown entity: translation along the $x$-axis. Therefore, to get the relative pose we use the following equation from the line intersection constraint
\begin{equation}
    m_1^T
    \begin{bmatrix}
    -[\mathbf{t}]_{\text{x}} \mathbf{R} & \mathbf{R} \\
    \mathbf{R} & \mathbf{0}
    \end{bmatrix}
    l_1 .
\end{equation}
We obtain a linear equation with a single unknown variable $t_1$. By applying the inverse transformations associated with the pre-processing ones, we can compute the relative pose necessary to align the two scans.

\subsection{Three Line Intersections and One Plane Correspondence (3L1P)} \label{sec:3l1p}
As shown in figure~\ref{fig:minimal_solvers}, we are given one plane correspondences $(\pi_1,\pi'_1)$ and three line intersection constraints $\{(l_1,m_1),(l_2,m_2),(l_3,m_3)\}$. The pre-processing transformation brings the two scans to the world coordinate frames such that the following condition is satisfied:
\begin{enumerate}
    \item The planes $\pi_1$ and $\pi'_1$ are on the $XY$ plane.
\end{enumerate}
By considering these assumptions, the relative pose between both frames is given by
\begin{equation}
    \label{eq:new_Rt}
    \mathbf{R} =
    \begin{bmatrix}
    c\theta & -s\theta & 0 \\
    s\theta & c\theta & 0 \\
    0 & 0 & 1 \\
    \end{bmatrix} \ \ \text{and} \
    \mathbf{t} =
    \begin{bmatrix}
    t_1\\t_2\\0
    \end{bmatrix},
\end{equation}
meaning that we reduced the total degrees of freedom from six to three\footnote{To simplify the notations, we use $c\theta = \text{cos}(\theta)$ and $s\theta = \text{sin}(\theta)$.}. To compute these unknowns, we use the three line intersections correspondences. We have three unknowns and three constraints from the line intersections as shown below:
\begin{equation}
    m_i^T
    \underbrace{\begin{bmatrix}
    -[\mathbf{t}]_{\text{x}} \mathbf{R} & \mathbf{R} \\
    \mathbf{R} & \mathbf{0}
    \end{bmatrix}}_{\mathbf{E}\in\mathbb{R}^{6\times 6}}
    l_i,i=\{1,2,3\}.
\end{equation}
After some algebraic manipulations (more detail is sent in the Supplementary Materials), we are able to get a single four degree polynomial as a function of the variable $s\theta$, which can be computed in closed-form. By solving this equation, we get $\theta$ and the remaining unknowns can be computed analytically, obtaining up to four possible solutions for $\mathbf{R}$ and $\mathbf{t}$. The optimal one is chosen in a RANSAC framework using additional correspondences. 
\section{Experiments}

\subsection{Datasets}

\noindent
\textbf{KITTI Dataset~\cite{Geiger2012CVPR}:} This consists of LiDAR point-clouds collected from the top of a moving vehicle. The LiDAR sensor captures roughly 10 fps (frames per second), with about 100k points per frame. 

\vspace{.25cm}
\noindent
\textbf{TUM Dataset~\cite{sturm12}:} This consists of sequences of Kinect RGBD data captured in an indoor environment. The sensor resolution is 640x480, at 30 fps. The sequences come with a ground truth trajectory of the sensor, obtained from a high-accuracy motion-capture system. Our algorithms only use the depth images from this dataset.

\subsection{Pre-processing}

\begin{figure}[ht]
    \centering
    \subfloat[original scan]{\includegraphics[width=0.5\linewidth]{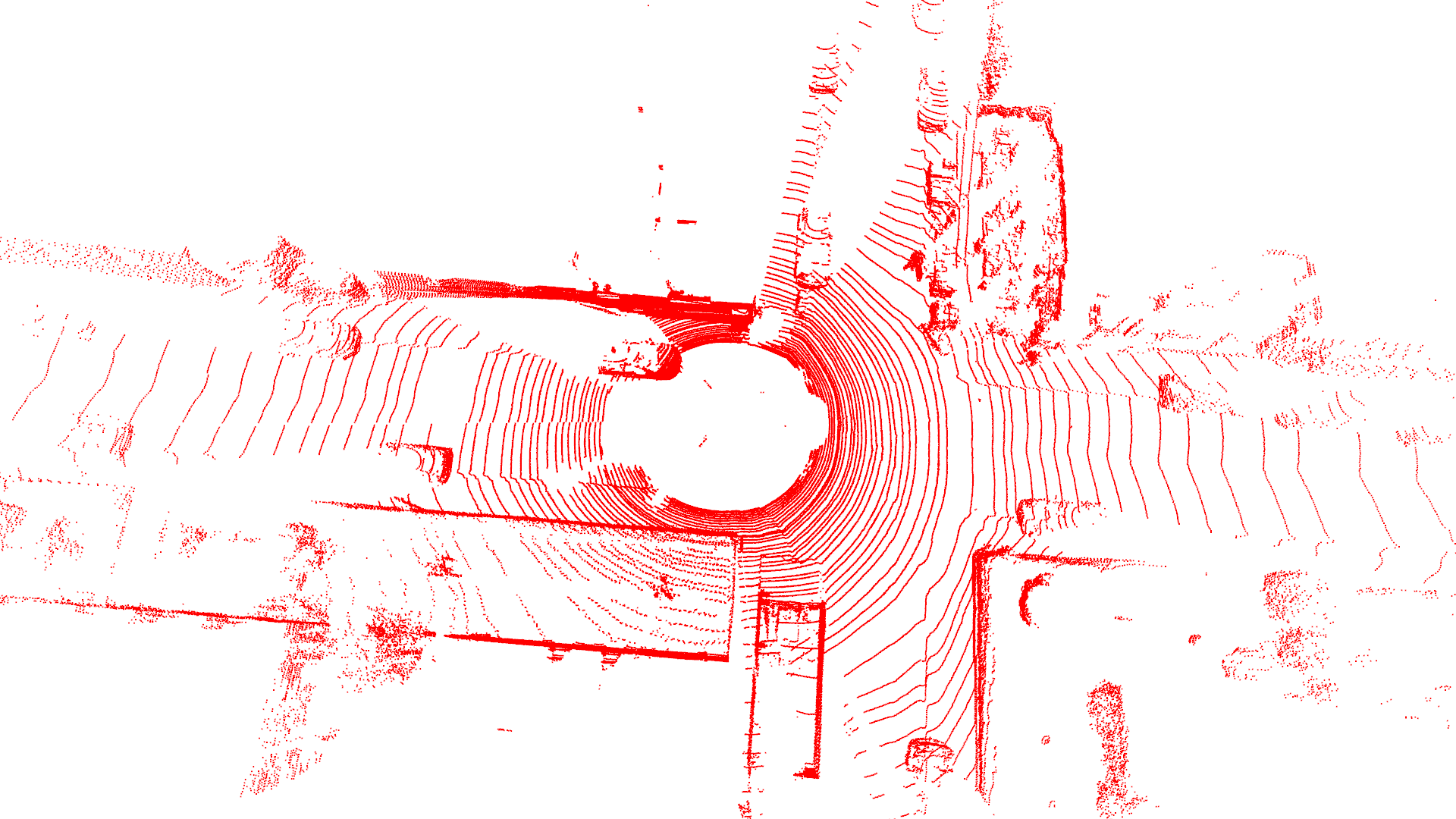}}
    \subfloat[1/36 downsampling]{\includegraphics[width=0.5\linewidth]{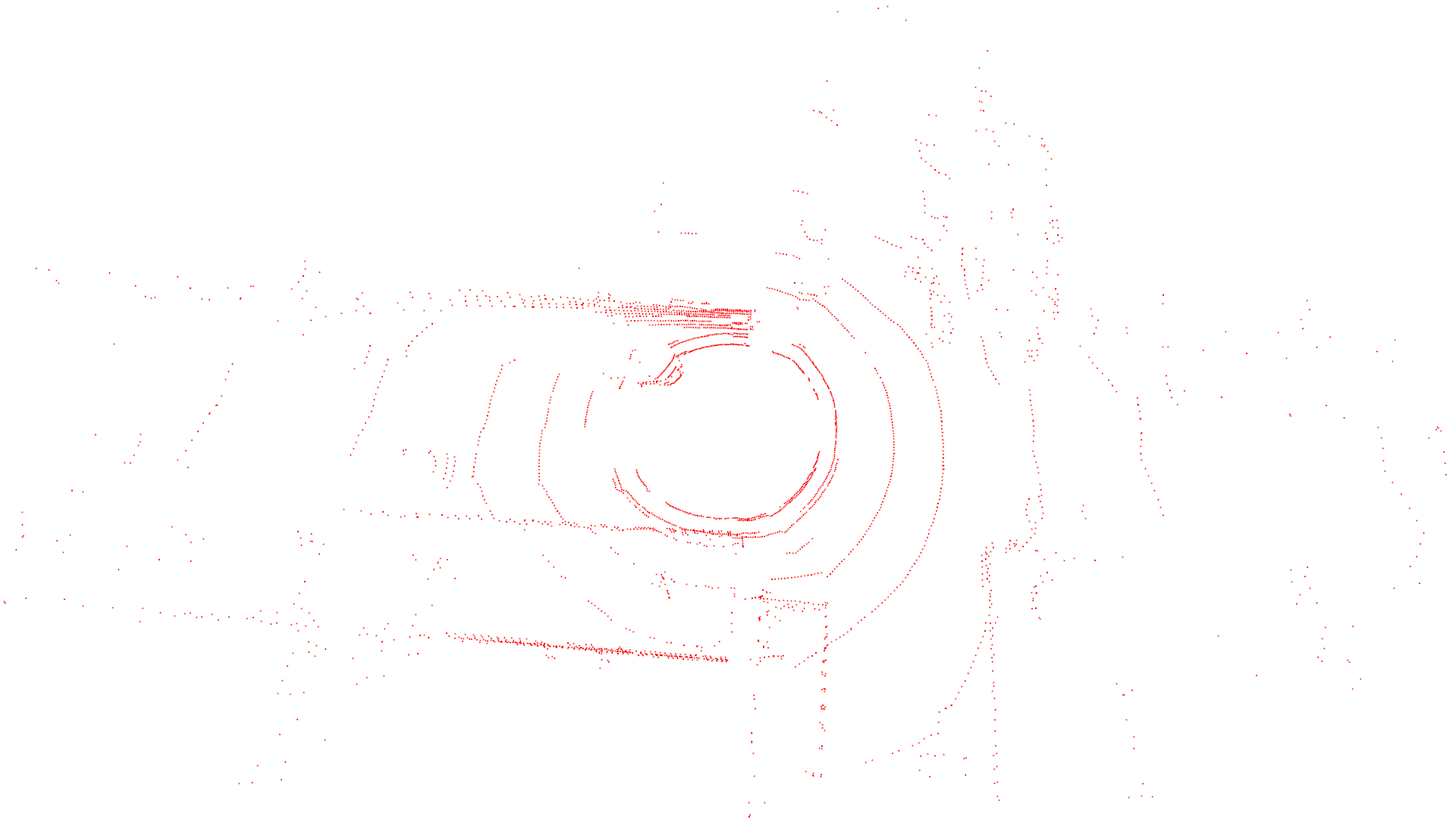}}
    \caption{Single frames from KITTI at (a) original resolution and (b) 1/36 down-sampled resolution}
    \label{fig:downsampling}
\end{figure}

\noindent
\textbf{Organized point-clouds:} Our algorithms require line and plane fitting outputs, which we find is easiest done on organized point-clouds. In the case of the KITTI data, we use the sensor calibration parameters to organize the raw input points into a grid-like structure by azimuth and elevation. In the case of Kinect data, the input is already the organized data, in the form of a depth image, and no further action is necessary. In experiments involving down-sampling of this data, we select points at appropriate indices from these organized point-clouds. For instance, figure~\ref{fig:downsampling} shows a point-cloud down-sampled by a factor of 6 along both horizontal and vertical directions, to retain roughly 1/36\textsuperscript{th} the points.

\begin{figure}[ht]
    \centering
    \includegraphics[width=\linewidth]{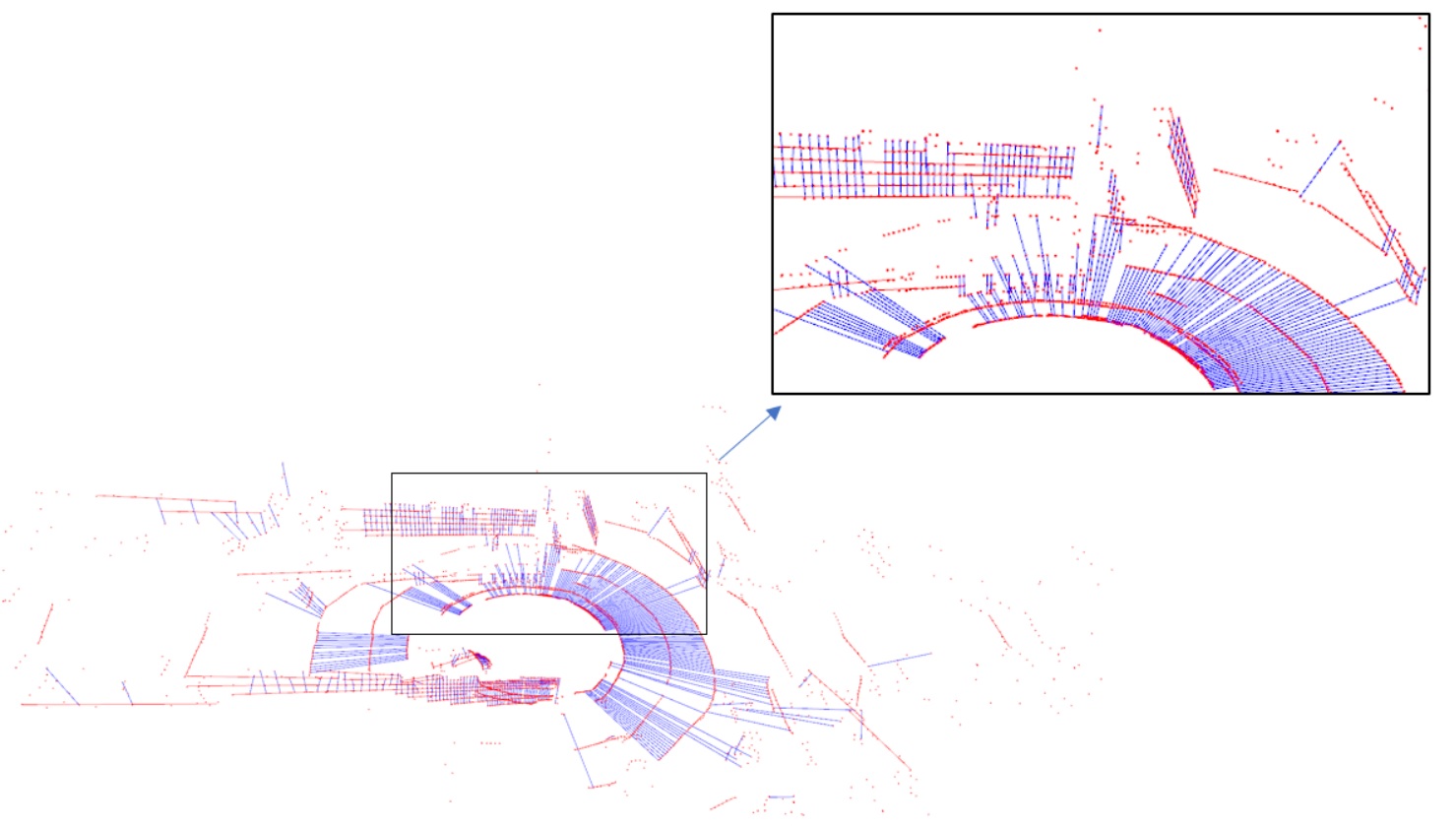}
    \caption{Visualization of line fitting on one KITTI frame. Red lines come from horizontal scan-lines and blue from vertical. We call them H-lines and V-lines respectively.}
    \label{fig:line-fiting}
\end{figure}

\vspace{.25cm}
\noindent
\textbf{Line Fitting:} We consider horizontal and vertical scan-lines in the organized point cloud and use RANSAC to do line fitting for every scan-line. We call lines that come from horizontal scan-lines ``H-lines'', and the ones coming from vertical scan lines are called ``V-lines''. For instance, in figure~\ref{fig:line-fiting}, H-lines are represented by the color red, and V-lines by blue. As we can see, H and V do not refer to the orientations of the lines in 3D, but the scan-line they come from. The blue lines on the ground plane in this figure are actually V-lines. Using intersections between these H- and V-lines (coming from the same frame), we can compute the normals for all lines, which we use in candidate selection step. All lines detected here come from some plane in the underlying scene, and these computed normal vectors are normals of those planes.

\begin{figure}[ht]
    \begin{center}
        \psfig{figure=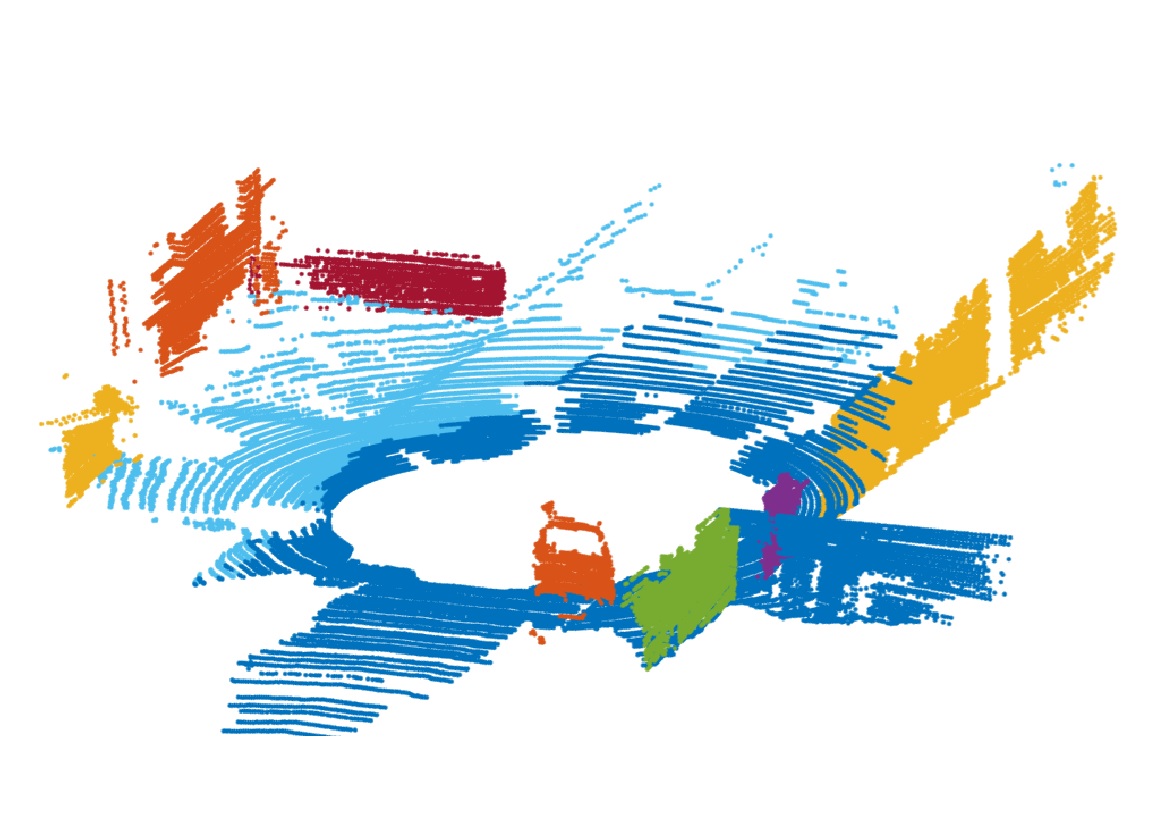,width=1.0\columnwidth}
    \end{center}
    \caption{Visualization of plane fitting on one KITTI frame. 10 planes are detected, with 8 above the ground.} 
    \label{fig:plane_fitting}
\end{figure}

\vspace{.25cm}
\noindent
\textbf{Plane Fitting:} We use RANSAC for plane fitting as well. 
KITTI has dense points on the ground. At least the ground plane will be detected, which is enough for the 3L1P method. 
Figure~\ref{fig:plane_fitting} shows a good example of plane fitting, with sufficiently many planes in front and at the back and on the sides of the street.

\subsection{Full Meta-Algorithm with RANSAC}

At each step, we estimate the frame-to-frame transformations between successive frames. If we designate the frame of the first camera position and orientation to be the world coordinate frame, we can multiply these transformations together to obtain transformation matrices for all frames.

All three of our solvers, 7 line intersections (7L), 1 line intersection with 2 plane correspondences (1L2P), and 3 line intersections with 1 plane correspondence (3L1P) produce a transformation matrix given a small number of inputs, with the latter two being minimal solvers. Thus, we apply these solvers in a RANSAC framework to obtain the best possible transformation matrices. 

\vspace{.25cm}
\noindent
\textbf{Candidates for RANSAC:} All of our solvers require some combination of (a) candidate line intersections from successive frames, and (b) candidate plane correspondences from successive frames. We get these candidates using the current estimate of the relative transformation between successive frames (initially the identity matrix).

\vspace{.25cm}
\noindent
\textbf{Candidate line intersections:} We consider all H-lines from one frame and V-lines from the other. Pairs of lines that are within a distance of some threshold (it is 2m for KITTI dataset) are considered to be candidate pairs.

\vspace{.25cm}
\noindent
\textbf{Candidate plane correspondences:} Among all the possible plane pairs from two frames, the candidate pairs have the angle of the plane normals smaller than 20 degree, and have the distance from the centroid of one plane to another plane within the same threshold used for candidate line intersection. 

\vspace{.25cm}
\noindent
\textbf{Candidate Selection:} For the algorithm 7L, it is necessary to select a good set of 7 line pairs in order to avoid degenerate cases. We cluster all the normal vectors from the first frame into 3 directions, and select 2 pairs uniformly at random from each cluster. The last pair is sampled (also uniform, random) from the full set of candidates, to get $2\times 3 + 1 = 7$ pairs. Similarly, for algorithm 3L1P, we sample one pair from each cluster. Plane correspondences are sampled uniformly at random from the full set of candidates.

\vspace{.25cm}
\noindent
\textbf{Inlier counting:} All three algorithms use line intersections to do inlier counting. From the set of candidates pairs, we count the (weighted) number of pairs that have a distance less than another threshold (2cm on KITTI dataset and 5mm on TUM here) after applying the computed transformation. Lines whose normals belong to clusters with a smaller number of lines get a higher weight, and vice versa.

In addition to the RANSAC, we find that in practice, we get better results by running the full algorithm described above thrice, each time using the best transformation from the previous step as the initial guess for candidate selection. As the candidates improve, so do the inliers.

\subsection{Evaluation Metrics}

\noindent
\textbf{Relative Pose Error (RPE):} As proposed in \cite{sturm12}, we compute the error in the relative pose between successive frames, and report the translation error in meters and rotation error in degrees respectively.

\vspace{.25cm}
\noindent
\textbf{Translational and Rotational Error along the trajectory:} This metric is used in \cite{Geiger2012CVPR} and on their online leaderboard. For all sub-sequences of length $100m, 200m, \ldots, 800m$, we compute the translation and rotation errors per unit length of the trajectory. Translation error is thus reported as a percentage value, and rotation error in degrees per meter.

\section{Results}
\begin{figure}[!h]
    \centering
    \subfloat[Sequence 5]{\includegraphics[width=0.33\linewidth]{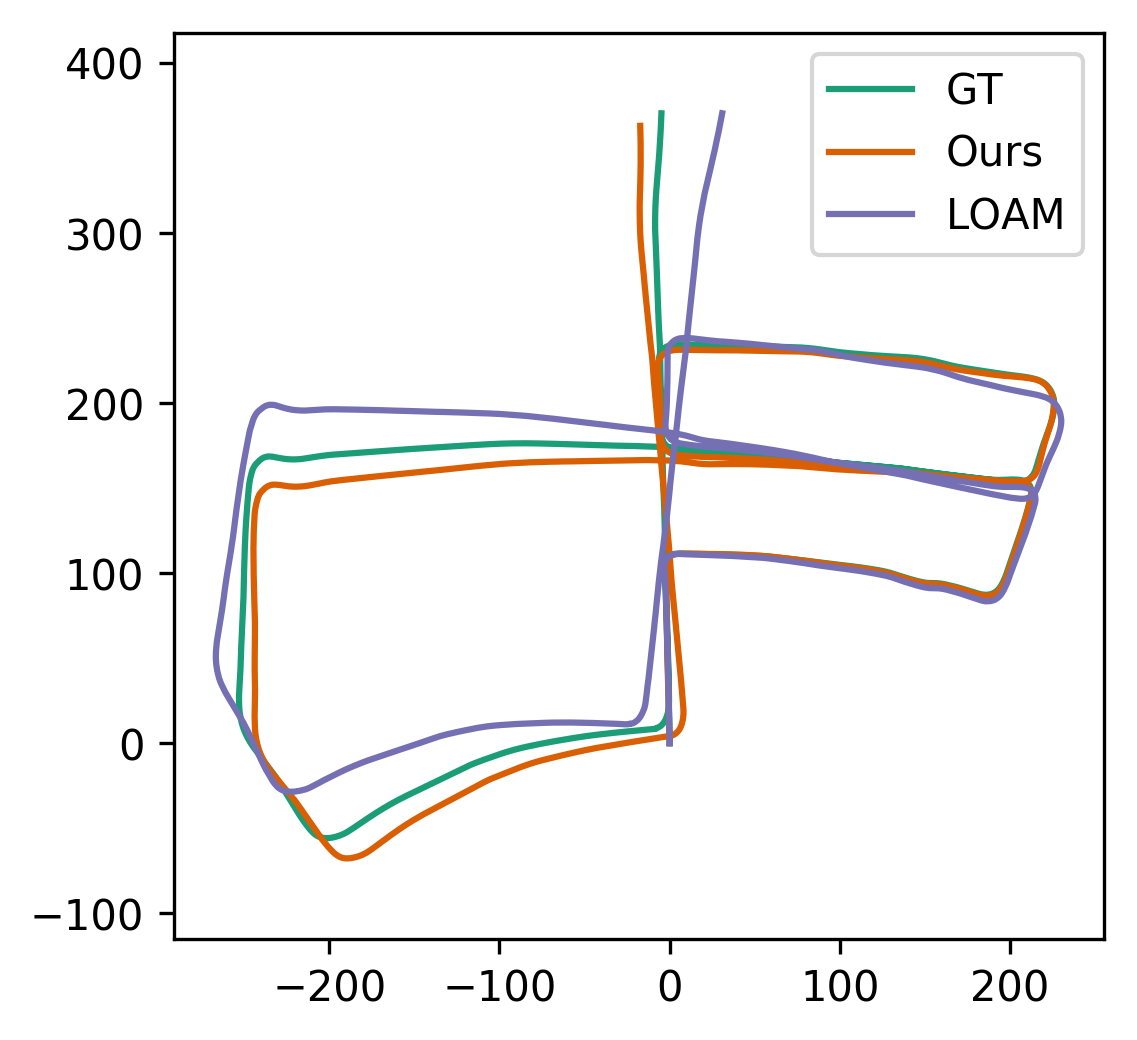}}
    \subfloat[Sequence 6]{\includegraphics[width=0.33\linewidth]{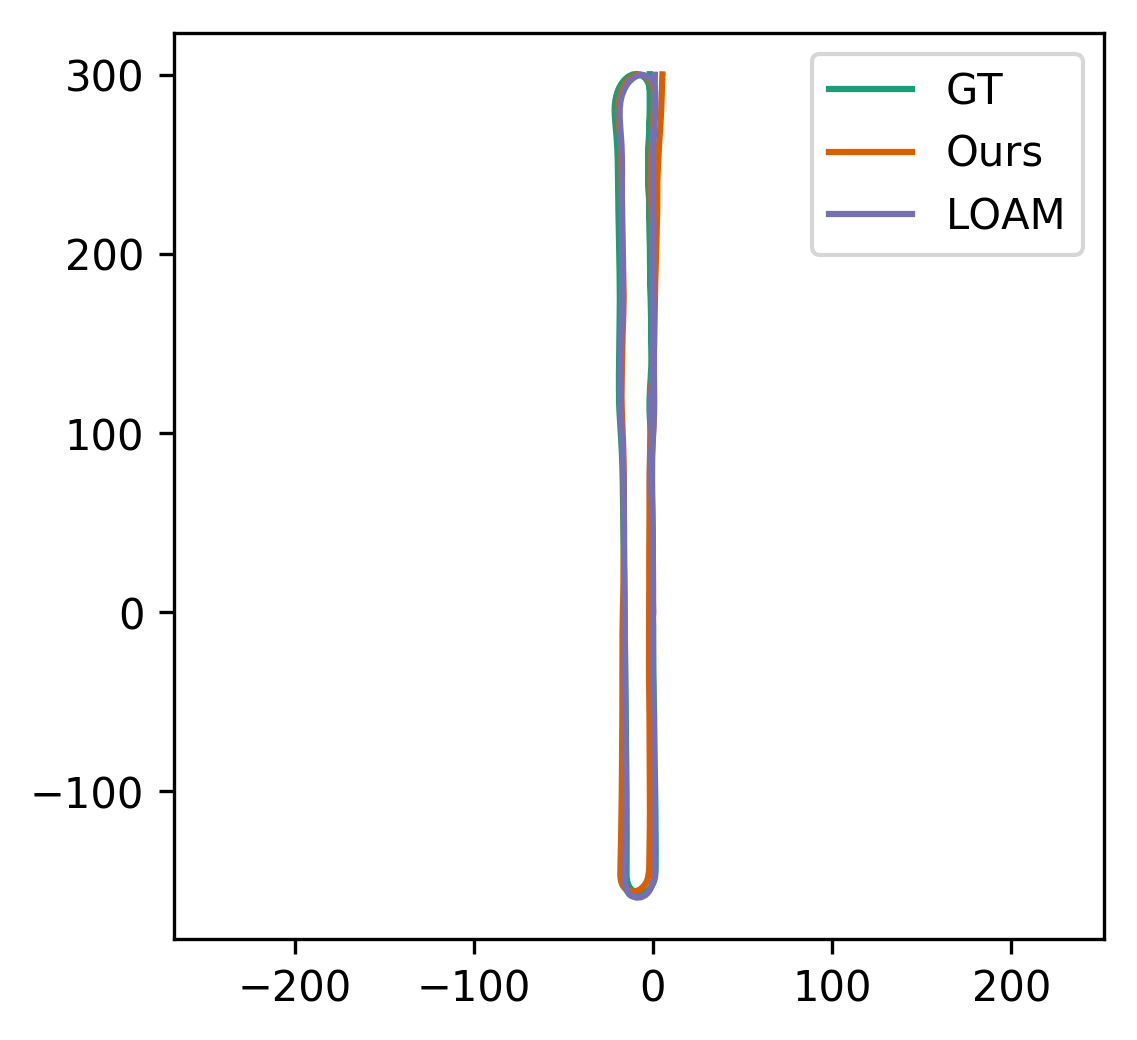}}
    \subfloat[Sequence 7]{\includegraphics[width=0.33\linewidth]{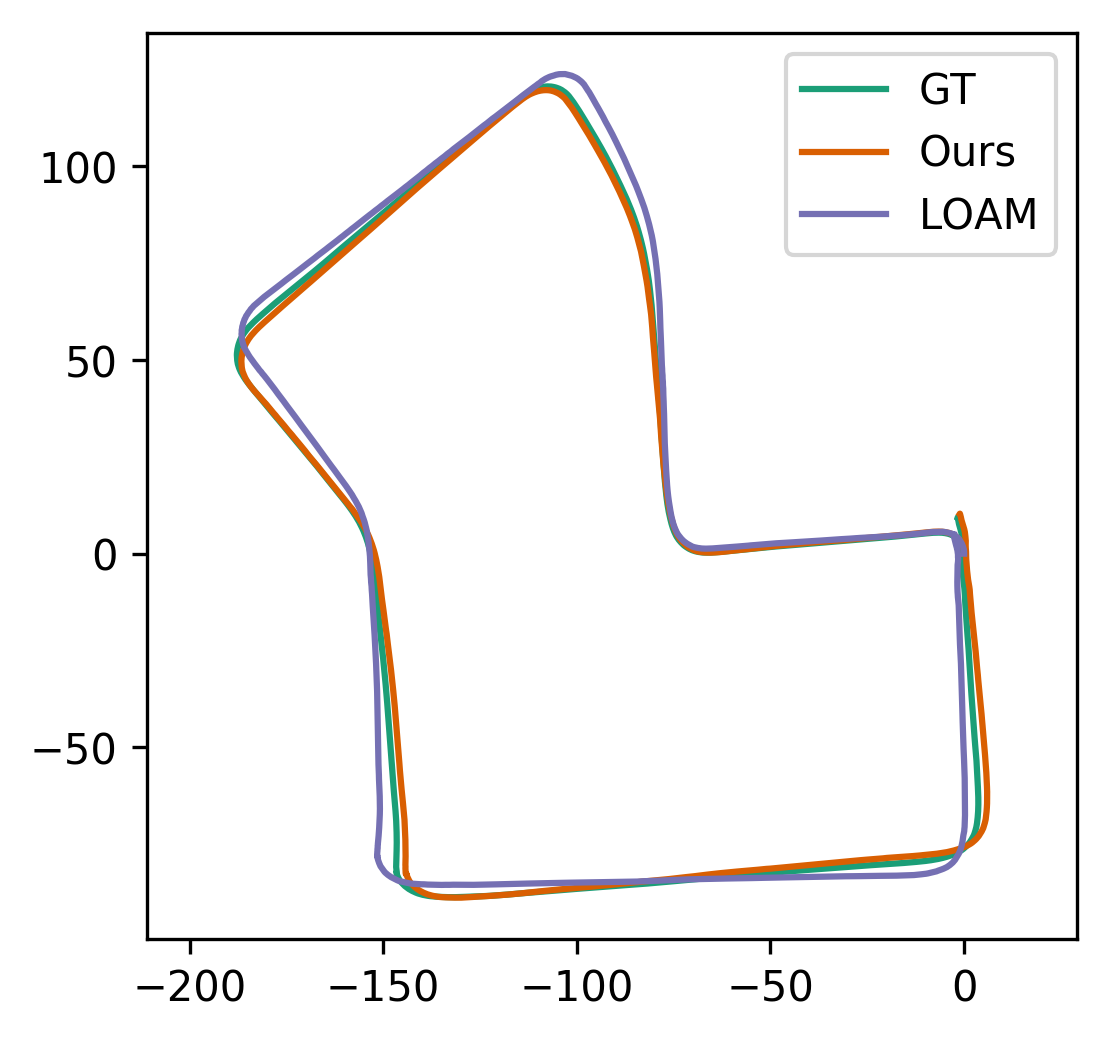}}
    \caption{Computed trajectories on KITTI Sequences at original resolution. The trajectory from the proposed method is closer to the ground truth than the one from LOAM.}
    \label{fig:kitti_trajectories}
\end{figure}

\begin{figure*}[!h]
    \centering
    \includegraphics[width=0.33\linewidth]{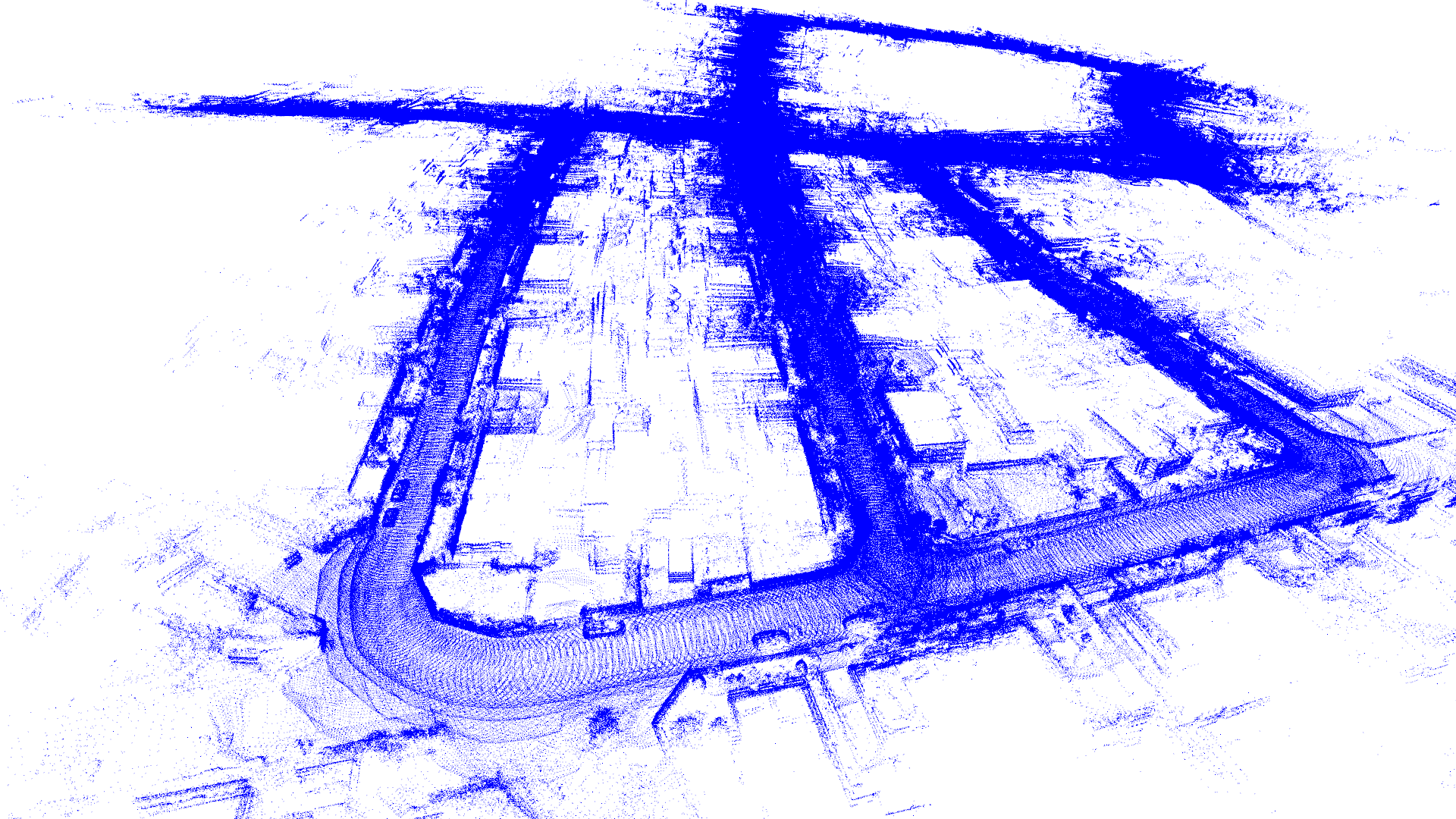}%
    \includegraphics[width=0.33\linewidth]{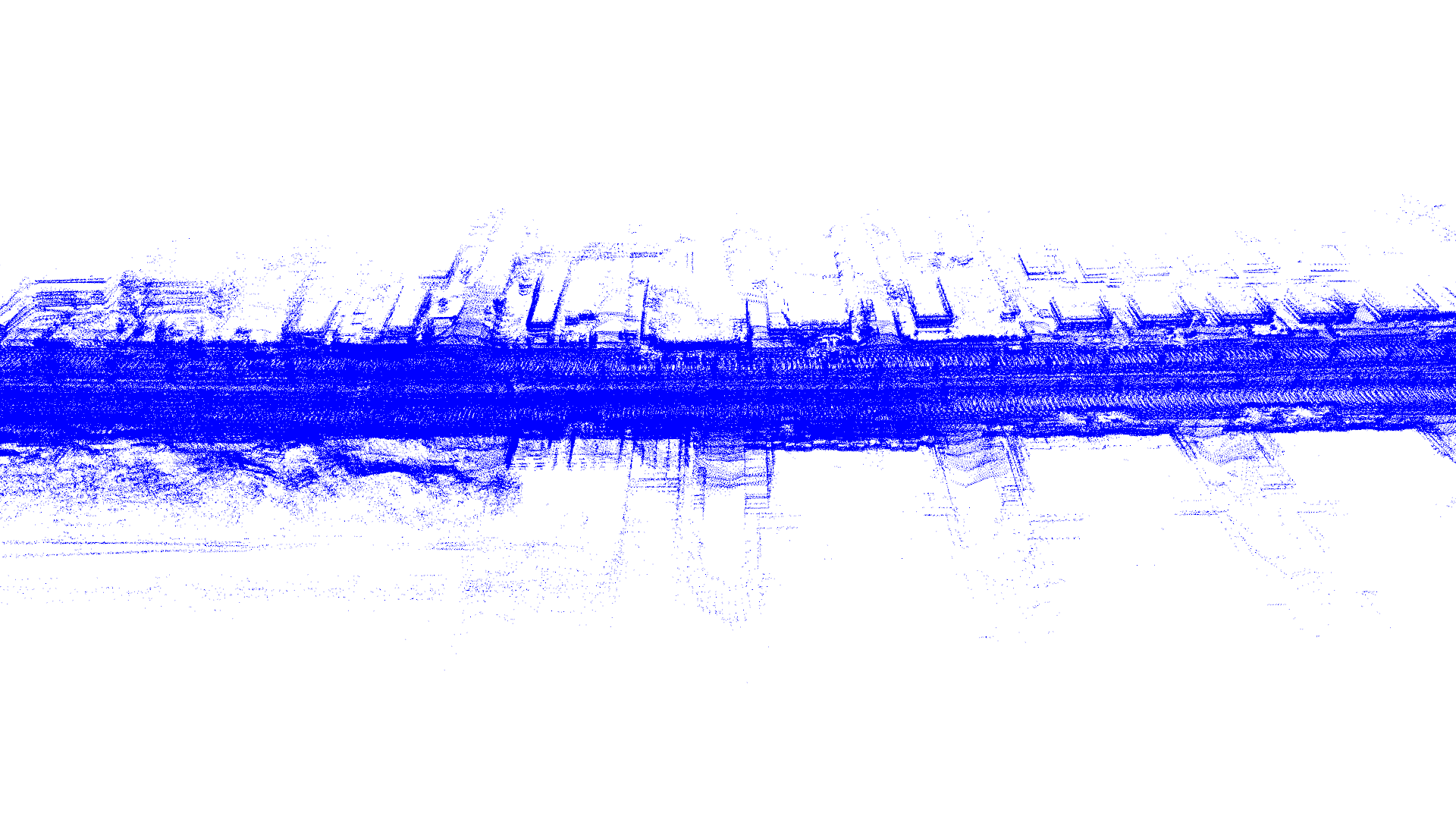}%
    \includegraphics[width=0.33\linewidth]{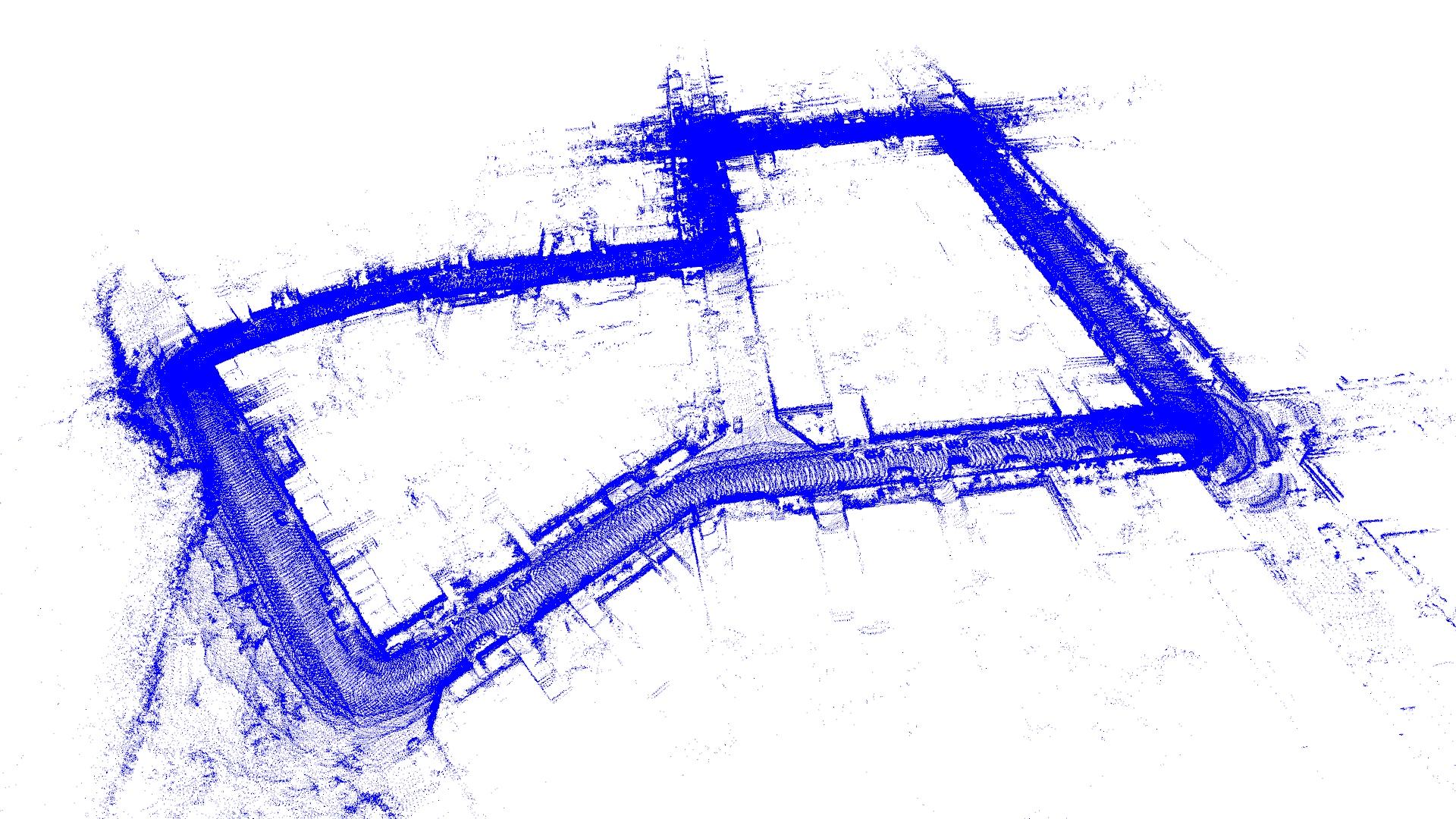}
    \caption{Visualization of the fully registered point-clouds from KITTI sequences 05, 06, and 07.}
    \label{fig:full_reg_kitti}
\end{figure*}

\noindent
{\bf LiDAR Data:}
We run our algorithm 7L on sequences 05, 06, and 07 of the KITTI dataset. We compare our results against the LOAM algorithm~\cite{Zhang14}, which is currently highly ranked on the online leaderboard. The computed trajectories on these sequences are shown in figure~\ref{fig:kitti_trajectories}, the relative pose error between consecutive frames is shown in table~\ref{tab:kitti_rpe}, and the error along the trajectory is reported in table~\ref{tab:kitti_trajectory_error}. The full point-clouds after registration are presented in figure~\ref{fig:full_reg_kitti}.

The results of the LOAM algorithm here might differ from those on the leaderboard because we run LOAM\footnote{we use the open-source version of LOAM available at \url{https://github.com/laboshinl/loam_velodyne}. The official version is no longer available.} by ourselves, without using IMU data, for a fair comparison.

\begin{table*}[ht]
    \centering
    \begin{tabular}{c|c|c|c|c|c|c|c|c}
        \multirow{2}{*}{Seq. ID} & \multicolumn{4}{c|}{Proposed} & \multicolumn{4}{c}{LOAM} \\ \cline{2-9}
        & \multicolumn{1}{P{0.08\linewidth}|}{mean tra. err. [m]} & \multicolumn{1}{P{0.08\linewidth}|}{mean rot. err. [deg]} & \multicolumn{1}{P{0.08\linewidth}|}{max tra. err. [m]} & \multicolumn{1}{P{0.08\linewidth}|}{max rot. err. [deg]} & \multicolumn{1}{P{0.08\linewidth}|}{mean tra. err. [m]} & \multicolumn{1}{P{0.08\linewidth}|}{mean rot. err. [deg]} & \multicolumn{1}{P{0.08\linewidth}|}{max tra. err. [m]} & \multicolumn{1}{P{0.08\linewidth}}{max rot. err. [deg]} \\ \hline
        05 & \textbf{0.0179} & \textbf{0.0955} & \textbf{0.1603} & \textbf{0.6806} & 0.0476 & 0.3120 & 1.1608 & 1.7840 \\
        06 & \textbf{0.0227} & \textbf{0.0792} & \textbf{0.0895} & \textbf{0.3139} & 0.0631 & 0.2111 & 1.2623 & 1.5848 \\
        07 & \textbf{0.0176} & \textbf{0.0838} & \textbf{0.0664} & \textbf{0.5690} & 0.0420 & 0.3214 & 0.6077 & 1.7964 \\
    \end{tabular}
    \caption{Mean and max errors between successive frames. Proposed method has a lower error than LOAM.}
    \label{tab:kitti_rpe}
\end{table*}

\begin{table}[ht]
    \centering
    \begin{tabular}{c|c|c|c|c}
        \multirow{2}{*}{Seq. ID} & \multicolumn{2}{c|}{Proposed} & \multicolumn{2}{c}{LOAM} \\ \cline{2-5}
        & \multicolumn{1}{P{0.15\linewidth}|}{tra. err. [\%]} & \multicolumn{1}{P{0.15\linewidth}|}{rot. err. [deg/m]} & \multicolumn{1}{P{0.15\linewidth}|}{tra. err. [\%]} & \multicolumn{1}{P{0.15\linewidth}}{rot. err. [deg/m]} \\ \hline
        05 & \textbf{1.4067} & \textbf{0.0088} & 1.7210 & 0.0110 \\
        06 & \textbf{0.8180} & \textbf{0.0056} & 1.8841 & 0.0097 \\
        07 & \textbf{0.9988} & \textbf{0.0063} & 2.0803 & 0.0138 \\
    \end{tabular}
    \caption{Results on 3 KITTI sequences. Error is measured as a fraction of the distance traveled on trajectory segments of lengths 100m, 200m, \ldots, 800, as specified in \cite{Geiger2012CVPR}.}
    \label{tab:kitti_trajectory_error}
\end{table}

We evaluate our 1L2P and 3L1P algorithms on KITTI sequence 07. The results for both error metrics are reported in table~\ref{tab:line_plane_result}.
\begin{table}[ht]
\centering
\begin{tabular}{P{1.5cm}|c|c|c|c|c}
\textbf{Metric} & \multicolumn{2}{c|}{\textbf{Successive Frames}} & \multicolumn{2}{c}{\textbf{KITTI}} \\ \hline
\multicolumn{1}{c|}{Method}     & \multicolumn{1}{c|}{\begin{tabular}[c]{@{}c@{}}tra. err.\\     {[}m{]}\end{tabular}} & \multicolumn{1}{c|}{\begin{tabular}[c]{@{}c@{}}rot. err\\ {[}deg{]}\end{tabular}} & \multicolumn{1}{c|}{\begin{tabular}[c]{@{}c@{}}tra. err.\\ {[}\%{]}\end{tabular}} & \multicolumn{1}{c}{\begin{tabular}[c]{@{}c@{}}rot. err.\\ {[}deg/m{]}\end{tabular}} \\ \hline
\multicolumn{1}{c|}{3L1P} & \multicolumn{1}{c|}{0.0769}                                                          & \multicolumn{1}{c|}{0.4473}                                                       & \multicolumn{1}{c|}{8.2284}                                                       & \multicolumn{1}{c}{0.0433}                                                          \\
\multicolumn{1}{c|}{1L2P} & \multicolumn{1}{c|}{0.1590}                                                          & \multicolumn{1}{c|}{0.8851}                                                       & \multicolumn{1}{c|}{28.3834}                                                      & \multicolumn{1}{c}{0.1813}                                                          \\
\end{tabular}
    \caption{Results of 3L1P and 1L2P on KITTI sequence 07}
    \label{tab:line_plane_result}
\end{table}

\begin{figure}[ht]
    \centering
    \subfloat[original data]{\includegraphics[width=0.33\linewidth]{figs/kitti_seq7_orig.png}}
    \subfloat[1/36 down-sampling]{\includegraphics[width=0.33\linewidth]{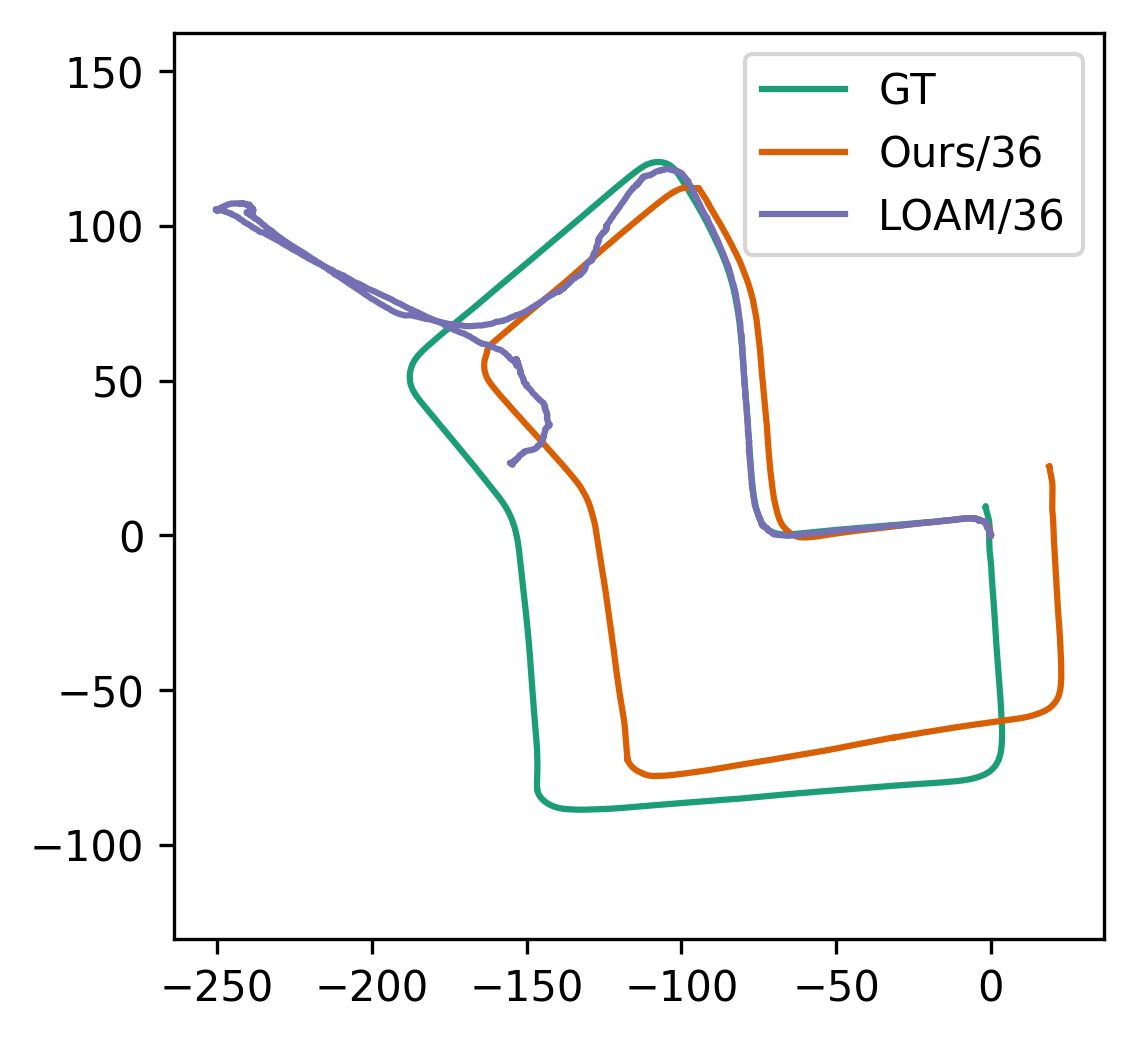}}
    \subfloat[1/64 down-sampling]{\includegraphics[width=0.33\linewidth]{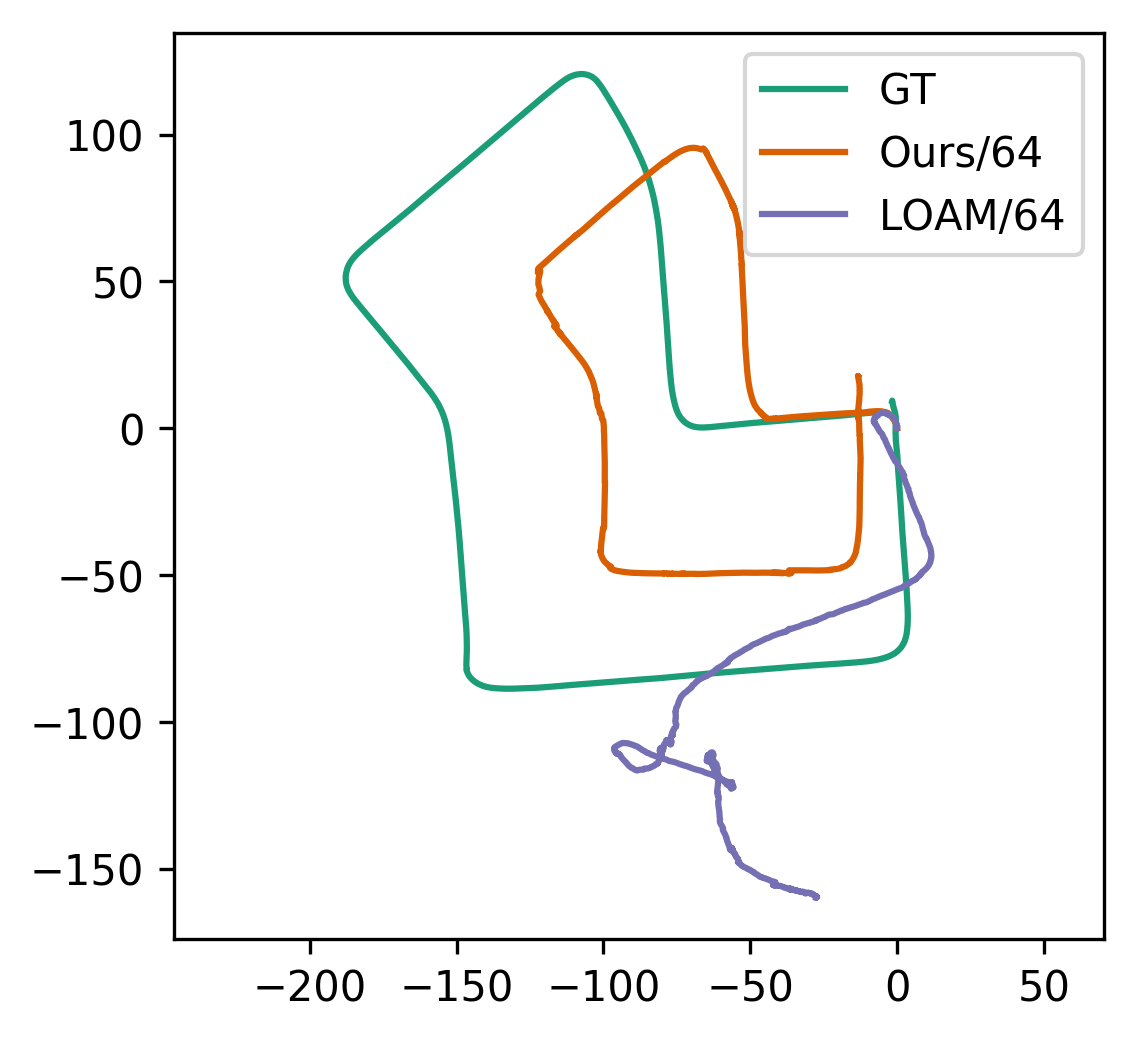}}
    \caption{Down-sampled KITTI sequences}
    \label{fig:kitti_07_down}
\end{figure}

We also evaluate our 7L method on down-sampled KITTI sequence 07. The trajectories are shown in figure~\ref{fig:kitti_07_down}, and the errors are reported in table~\ref{tab:kitti_07_down}. As can be seen in the trajectory (and indeed the error values), our 7L algorithm outperforms LOAM on sparse data.

\begin{table}[ht]
    \centering
    \begin{tabular}{P{1.5cm}|c|c|c|c}
        \multirow{2}{\linewidth}{\parbox{1.5cm}{\centering Down-sampling factor}} & \multicolumn{2}{c|}{Proposed} & \multicolumn{2}{c}{LOAM} \\ \cline{2-5}
        & \multicolumn{1}{P{0.15\linewidth}|}{tra. err. [\%]} & \multicolumn{1}{P{0.15\linewidth}|}{rot. err. [deg/m]} & \multicolumn{1}{P{0.15\linewidth}|}{tra. err. [\%]} & \multicolumn{1}{P{0.15\linewidth}}{rot. err. [deg/m]} \\ \hline%
        36 & \textbf{7.4192} &\textbf{0.0234} & 52.2320 & 0.3485 \\
        64 & \textbf{23.1660} & \textbf{0.1065} & 58.6810 & 0.4940 \\
    \end{tabular}
    \caption{Results on down-sampled KITTI sequence 07}
    \label{tab:kitti_07_down}
\end{table}


\vspace{.25cm}
\noindent
{\bf Kinect Data:} We test our 7L algorithm on 3 sequences from the TUM dataset, down-sampled by a factor of 10 in both dimensions (i.e.\ 1/100\textsuperscript{th} the points). The error values for all three sequences are presented in table~\ref{tab:TUM}. Our method outperforms the standard 3-point pose estimation with RANSAC baseline (which uses the RGB, and runs at the original resolution) in spite of us not using the RGB data at all, and using only 1/100\textsuperscript{th} the points.

\begin{table}[!h]
    \centering
    \begin{tabular}{c|c|c}
        Seq. ID & \multicolumn{1}{P{0.15\linewidth}|}{tra. err. [mm]} & \multicolumn{1}{P{0.15\linewidth}}{rot. err. [deg]} \\ \hline
        \texttt{fr1/xyz} & \textbf{5.047} & \textbf{0.5212} \\
        \texttt{fr3/sitting\_xyz} & \textbf{11.633} & \textbf{0.6913} \\
        \texttt{fr1/room} & \textbf{12.043} & \textbf{0.6949} \\
    \end{tabular}
    \caption{Results on 3 sequences of the TUM RGBD dataset. We use only the depth-maps, down-sampled by a factor of 100. Mean rotation and translation error between successive frames is reported. Standard 3-point solvers in RANSAC produces an average error of 0.740 for rotation, and 24 mm for translation on the original resolution with RGB data.}
    \label{tab:TUM}
\end{table}

\vspace{.25cm}
\noindent
{\bf Performance and Speed:} All of our solvers are implemented in C/C++. The timings for various operations (running on 1 core of Intel i7-8700K) are given in table~\ref{tab:speed}. The time taken by inlier counting varies with the number of detected lines; table contains typically observed values.

\begin{table}[!h]
    \centering
    \scalebox{.8}{
    \begin{tabular}{P{.16\linewidth}|P{.16\linewidth}|P{.16\linewidth}|P{.16\linewidth}|P{.16\linewidth}|P{.16\linewidth}}
         & 7 Lines & 3 Lines \& 1 Plane & 1 Line \& 2 Plane & Inlier counting & line fitting \\ \hline
        Time [$\mu s$] & 22k (=0.022 s) &  4.9 & 0.64 & 10 to 2000 & 3M (=3s) \\
    \end{tabular}}
    \caption{Average computation time for each of the proposed solvers, and other tasks.}
    \label{tab:speed}
\end{table}

We use a threshold (2 cm on KITTI, 5 mm on TUM) as well as max number of iterations (30K iterations) for terminating the alternating projection algorithm. In the future, we will explore certain extrapolated projection techniques for further speedup~\cite{Censor2012}.


\section{Discussion}\label{sec:discussion}

The proposed algorithm is applicable to sensors on a moving platform and we assume smoothness assumption for obtaining line intersection constraints, although our algorithm is robust to outliers due to the use of minimal solvers in a RANSAC framework. 

While we outperform LOAM on many settings, our method can further be improved by correcting for distortions obtained from moving platforms, and utilizing lines or edge points from boundary areas, as it is done in LOAM~\cite{Zhang14}. 

In our work we observed that alternating minimization can be used for solving near-minimal problems that are known to be hard using algebraic solvers. This raises an important question of whether alternating projection techniques can be extended to other hard minimal problems.     

{\small
\bibliographystyle{ieee}
\bibliography{main}

\begin{thebibliography}{10}\itemsep=-1pt

\bibitem{arun87}
K.~S. {Arun}, T.~S. {Huang}, and S.~D. {Blostein}.
\newblock Least-squares fitting of two 3-d point sets.
\newblock {\em IEEE Trans. Pattern Analysis and Machine Intelligence (T-PAMI)},
  9(5):698--700, 1987.

\bibitem{besl92}
P.~Besl and N.~McKay.
\newblock A method for registration of 3-{D} shapes.
\newblock {\em IEEE Trans. Pattern Analysis and Machine Intelligence (T-PAMI)},
  14(2):239--256, 1992.

\bibitem{bhattacharya17}
U.~Bhattacharya, S.~Veerawal, and V.~M. Govindu.
\newblock Fast multiview {3D} scan registration using planar structures.
\newblock In {\em Int'l Conf. on 3D Vision (3DV)}, pages 548--556, 2017.

\bibitem{Campbell16}
D.~{Campbell} and L.~{Petersson}.
\newblock {GOGMA}: {G}lobally-optimal gaussian mixture alignment.
\newblock In {\em IEEE Conf. Computer Vision and Pattern Recognition (CVPR)},
  pages 5685--5694, 2016.

\bibitem{camposeco18}
F.~Camposeco, A.~Cohen, M.~Pollefeys, and T.~Sattler.
\newblock Hybrid camera pose estimation.
\newblock In {\em IEEE Conf. Computer Vision and Pattern Recognition (CVPR)},
  pages 136--144, 2018.

\bibitem{camposeco16}
F.~Camposeco, T.~Sattler, and M.~Pollefeys.
\newblock Minimal solvers for generalized pose and scale estimation from two
  rays and one point.
\newblock In {\em European Conf. on Computer Vision (ECCV)}, pages 202--218,
  2016.

\bibitem{Censor2012}
Y.~Censor, W.~Chen, P.~Combettes, R.~Davidi, and G.~Herman.
\newblock On the effectiveness of projection methods for convex feasibility
  problems with linear inequality constraints.
\newblock {\em Computational Optimization and Applications}, 2012.

\bibitem{choi13}
C.~Choi, A.~J.~B. Trevor, and H.~I. Christensen.
\newblock Rgb-d edge detection and edge-based registration.
\newblock In {\em IEEE/RSJ Int'l Conf. on Intelligent Robots and Systems
  (IROS)}, pages 1568--1575, 2013.

\bibitem{Deschaud18}
J.~{Deschaud}.
\newblock {IMLS-SLAM}: {S}can-to-model matching based on {3D} data.
\newblock In {\em IEEE Int'l Conf. Robotics and Automation (ICRA)}, pages
  2480--2485, 2018.

\bibitem{Ding2019}
L.~Ding and C.~Feng.
\newblock {DeepMapping}: {U}nsupervised map estimation from multiple point
  clouds.
\newblock In {\em IEEE Conf. Computer Vision and Pattern Recognition (CVPR)},
  2019.

\bibitem{elbaz17}
G.~Elbaz, T.~Avraham, and A.~Fischer.
\newblock {3D} point cloud registration for localization using a deep neural
  network auto-encoder.
\newblock In {\em IEEE Conf. Computer Vision and Pattern Recognition (CVPR)},
  pages 2472--2481, 2017.

\bibitem{Elbaz2017}
G.~Elbaz, T.~Avraham, and A.~Fischer.
\newblock {3D Point Cloud Registration for Localization using a Deep Neural
  Network Auto-Encoder}.
\newblock In {\em IEEE Conf. Computer Vision and Pattern Recognition (CVPR)},
  pages 2472 -- 2481, 2017.

\bibitem{endres12}
F.~Endres, J.~Hess, N.~Engelhard, J.~Sturm, D.~Cremers, and W.~Burgard.
\newblock An evaluation of the rgb-d slam system.
\newblock In {\em IEEE Int'l Conf. on Robotics and Automation (ICRA)}, pages
  1691--1696, 2012.

\bibitem{endres14}
F.~Endres, J.~Hess, J.~Sturm, D.~Cremers, and W.~Burgard.
\newblock {3-D} mapping with an {RGB-D} camera.
\newblock {\em IEEE Trans. on Robotics (T-Ro)}, 30(1):177--187, 2014.

\bibitem{fischler81}
M.~A. Fischler and R.~C. Bolles.
\newblock {R}andom {S}ample {C}onsensus: {A} paradigm for model fitting with
  applications to image analysis and automated cartography.
\newblock {\em Commun. ACM}, 24(6):381--395, 1981.

\bibitem{fraundorfer10}
F.~Fraundorfer, P.~Tanskanen, and M.~Pollefeys.
\newblock A minimal case solution to the calibrated relative pose problem for
  the case of two known orientation angles.
\newblock In {\em European Conf. on Computer Vision (ECCV)}, pages 269--282,
  2010.

\bibitem{Geiger2012CVPR}
A.~Geiger, P.~Lenz, and R.~Urtasun.
\newblock Are we ready for autonomous driving? the kitti vision benchmark
  suite.
\newblock In {\em Conference on Computer Vision and Pattern Recognition
  (CVPR)}, 2012.

\bibitem{Grant2018}
W.~S. Grant, R.~C. Voorhies, and L.~Itti.
\newblock Efficient velodyne {SLAM} with point and plane features.
\newblock {\em Autonomous Robots}, 2018.

\bibitem{grossberg01}
M.~Grossberg and S.~Nayar.
\newblock A general imaging model and a method for finding its parameters.
\newblock In {\em ICCV}, 2001.

\bibitem{horn87}
B.~K.~P. Horn.
\newblock Closed-form solution of absolute orientation using unit quaternions.
\newblock {\em Journal of the Optical Society of America A}, 4(4):629--642,
  1987.

\bibitem{ke17}
T.~Ke and S.~I. Roumeliotis.
\newblock An efﬁcient algebraic solution to the perspective-three-point
  problem.
\newblock In {\em IEEE Conf. on Computer Vision and Pattern Recognition
  (CVPR)}, pages 4618--4626, 2017.

\bibitem{khoury17}
M.~Khoury, Q.~Zhou, and V.~Koltun.
\newblock Learning compact geometric features.
\newblock In {\em IEEE Int'l Conf. on Computer Vision (ICCV)}, pages 153--161,
  2017.

\bibitem{kneip11}
L.~Kneip, D.~Scaramuzza, and R.~Siegwart.
\newblock A novel parametrization of the perspective-three-point problem for a
  direct computation of absolute camera position and orientation.
\newblock In {\em IEEE Conf. on Computer Vision and Patter Recognition (CVPR)},
  pages 2969--2976, 2011.

\bibitem{kneip12}
L.~Kneip, R.~Siegwart, and M.~Pollefeys.
\newblock Finding the exact rotation between two images independently of the
  translation.
\newblock In {\em European Conf. Computer Vision (ECCV)}, pages 696--709, 2012.

\bibitem{li13}
B.~Li, L.~Heng, G.~H. Lee, and M.~Pollefeys.
\newblock A 4-point algorithm for relative pose estimation of a calibrated
  camera with a known relative rotation angle.
\newblock In {\em IEEE/RSJ Int'l Conf. on Intelligent Robots and Systems
  (IROS)}, pages 1595--1601, 2013.

\bibitem{li06}
H.~Li.
\newblock A simple solution to the six-point two-view focal-length problem.
\newblock In {\em European Conf. Computer Vision (ECCV)}, pages 200--213, 2006.

\bibitem{li06b}
H.~Li and R.~Hartley.
\newblock Five-point motion estimation made easy.
\newblock In {\em Int'l Conf. on Pattern Recognition (ICPR)}, volume~1, pages
  630--633, 2006.

\bibitem{li17}
H.~Li and R.~Hartley.
\newblock The {3D-3D} registration problem revisited.
\newblock In {\em IEEE Int'l Conf. Computer Vision (ICCV)}, pages 1--8, 2017.

\bibitem{Li2008}
H.~Li, R.~Hartley, and J.~Kim.
\newblock A linear approach to motion estimation using generalized camera
  models.
\newblock In {\em CVPR}, 2008.

\bibitem{liu18}
C.~Liu, J.~Wu, and Y.~Furukawa.
\newblock {FloorNet}: {A} unified framework for floorplan reconstruction from
  3d scans.
\newblock In {\em European Conf. on Computer Vision (ECCV)}, 2018.

\bibitem{Liu18_eccv}
Y.~Liu, C.~Wang, Z.~Song, and M.~Wang.
\newblock Efficient global point cloud registration by matching rotation
  invariant features through translation search.
\newblock In {\em European Conf. Computer Vision (ECCV)}, pages 460--474, 2018.

\bibitem{lu15}
Y.~Lu and D.~Song.
\newblock Robust {RGB-D} odometry using point and line features.
\newblock In {\em IEEE Int'l Conf. on Computer Vision (ICCV)}, pages
  3934--3942, 2015.

\bibitem{ma16}
L.~Ma, C.~Kerl, J.~St\"uckler, and D.~Cremers.
\newblock {CPA-SLAM:} {C}onsistent plane-model alignment for direct {RGB-D
  SLAM}.
\newblock In {\em IEEE Int'l Conf. on Robotics and Automation (ICRA)}, pages
  1285--1291, 2016.

\bibitem{Makadia06_cvpr}
A.~{Makadia}, A.~{Patterson}, and K.~{Daniilidis}.
\newblock Fully automatic registration of 3d point clouds.
\newblock In {\em IEEE Conf. Computer Vision and Pattern Recognition (CVPR)},
  volume~1, pages 1297--1304, 2006.

\bibitem{nister03}
D.~Nist\'{e}r.
\newblock An efficient solution to the five-point relative pose problem.
\newblock In {\em Conference on Computer Vision and Pattern Recognition
  (CVPR)}, 2003.

\bibitem{penney01}
G.~P. Penney, P.~J. Edwards, A.~P. King, J.~M. Blackall, P.~G. Batchelor, and
  D.~J. Hawkes.
\newblock A stochastic iterative closest point algorithm (stochast{ICP}).
\newblock In {\em Medical Image Computing and Computer-Assisted Intervention
  (MICCAI)}, pages 762--769, 2001.

\bibitem{persson18}
M.~Persson and K.~Nordberg.
\newblock {L}ambda twist: {A}n accurate fast robust perspective three point
  ({P3P}) solver.
\newblock In {\em European Conf. on Computer Vision (ECCV)}, pages 334--349,
  2018.

\bibitem{pless03}
R.~Pless.
\newblock Using many cameras as one.
\newblock In {\em IEEE Conf. Computer Vision and Pattern Recognition (CVPR)},
  volume~2, page 587, June 2003.

\bibitem{ROB-035}
F.~Pomerleau, F.~Colas, and R.~Siegwart.
\newblock A review of point cloud registration algorithms for mobile robotics.
\newblock {\em Foundations and Trends® in Robotics}, 4(1):1--104, 2015.

\bibitem{pottmann01}
H.~Pottmann and J.~Wallner.
\newblock {\em Computational Line Geometry}.
\newblock Springer-Verlag Berlin Heidelberg, 1 edition, 2001.

\bibitem{saurer15}
O.~Saurer, P.~Vasseur, C.~Demonceaux, and F.~Fraundorfer.
\newblock A homography formulation to the 3pt plus a common direction relative
  pose problem.
\newblock In {\em Asian Conf. on Computer Vision (ACCV)}, pages 288--301, 2015.

\bibitem{schonemann66}
P.~H. Sch\"{o}nemann.
\newblock A generalized solution of the orthogonal procrustes problem.
\newblock {\em Psychometrika}, 31(1):1--10, 1966.

\bibitem{stewenius05b}
H.~Stewenius, D.~Nister, F.~Kahl, and F.~Schaffalitzky.
\newblock A minimal solution for relative pose with unknown focal length.
\newblock In {\em IEEE Conf. on Computer Vision and Pattern Recognition
  (CVPR)}, volume~2, pages 789--794, 2005.

\bibitem{stewenius05}
H.~Stewenius, D.~Nister, and K.~A. M.~Oskarsson.
\newblock Solutions to minimal generalized relative pose problems.
\newblock In {\em Workshop on Omnidirectional Vision (OMNIVIS)}, 2005.

\bibitem{Stewenius2005}
H.~Stewenius, D.~Nister, M.~Oskarsson, and K.~Astrom.
\newblock Solutions to minimal generalized relative pose problems.
\newblock In {\em OMNIVIS}, 2005.

\bibitem{stewenius05c}
H.~Stewenius, D.~Nister, M.~Oskarsson, and K.~Astrom.
\newblock Solutions to minimal generalized relative pose problems.
\newblock In {\em Workshop on Omnidirectional Vision (OMNIVIS)}, 2005.

\bibitem{Straub17_cvpr}
J.~Straub, T.~Campbell, J.~How, and J.~W. Fisher.
\newblock Efficient global point cloud alignment using bayesian nonparametric
  mixtures.
\newblock In {\em IEEE Conf. Computer Vision and Pattern Recognition (CVPR)},
  pages 2403--2412, 2017.

\bibitem{sturm12}
J.~Sturm, N.~Engelhard, F.~Endres, W.~Burgard, and D.~Cremers.
\newblock A benchmark for the evaluation of {RGB-D SLAM} systems.
\newblock In {\em IEEE/RSJ Int'l Conf. on Intelligent Robots and Systems
  (IROS)}, pages 573--580, 2012.

\bibitem{sturm05}
P.~Sturm.
\newblock Multi-view geometry for general camera models.
\newblock In {\em CVPR}, 2005.

\bibitem{THEILER2015126}
P.~W. Theiler, J.~D. Wegner, and K.~Schindler.
\newblock Globally consistent registration of terrestrial laser scans via graph
  optimization.
\newblock {\em ISPRS Journal of Photogrammetry and Remote Sensing},
  109:126--138, 2015.

\bibitem{umeyama91}
S.~Umeyama.
\newblock Least-squares estimation of transformation parameters between two
  point patterns.
\newblock {\em IEEE Trans. Pattern Analysis and Machine Intelligence (T-PAMI)},
  13(4):376--380, 1991.

\bibitem{ventura15}
J.~Ventura, C.~Arth, and V.~Lepetit.
\newblock An efficient minimal solution for multi-camera motion.
\newblock In {\em IEEE Int'l Conf. on Computer Vision (ICCV)}, pages 747--755,
  2015.

\bibitem{ventura14}
J.~Ventura, C.~Arth, G.~Reitmayr, and D.~Schmalstieg.
\newblock A minimal solution to the generalized pose-and-scale problem.
\newblock In {\em IEEE Conf. on Computer Vision and Pattern Recognition
  (CVPR)}, pages 422--429, 2014.

\bibitem{wang18}
P.~Wang, G.~Xu, Z.~Wang, and Y.~Cheng.
\newblock An efficient solution to the perspective-three-point pose problem.
\newblock {\em Computer Vision and Image Understanding (CVIU)}, 166:81--87,
  2018.

\bibitem{yan15}
J.~Yan, J.~Wang, H.~Zha, X.~Yang, and S.~Chu.
\newblock Multi-view point registration via alternating optimization.
\newblock In {\em AAAI Conference on Artificial Intelligence}, 2015.

\bibitem{yang16}
J.~Yang, H.~Li, D.~Campbell, and Y.~Jia.
\newblock {Go-ICP}: {A} globally optimal solution to 3{D} {ICP} point-set
  registration.
\newblock {\em IEEE Trans. Pattern Analysis and Machine Intelligence (T-PAMI)},
  38(11):2241--2254, 2016.

\bibitem{yang13}
J.~Yang, H.~Li, and Y.~Jia.
\newblock {Go-ICP}: {S}olving 3d registration efficiently and globally
  optimally.
\newblock In {\em IEEE Int'l Conf. Computer Vision (ICCV)}, pages 1457--1464,
  2013.

\bibitem{Zhang14}
J.~Zhang and S.~Singh.
\newblock {LOAM}: {L}idar odometry and mapping in real-time.
\newblock In {\em Robotics: Science and Systems (RSS)}, 2014.

\bibitem{Zhang17}
J.~Zhang and S.~Singh.
\newblock Low-drift and real-time lidar odometry and mapping.
\newblock {\em Autonomous Robots}, 41(2):401--416, 2017.

\bibitem{Zhou-2018-107715}
L.~Zhou, Z.~Li, and M.~Kaess.
\newblock Automatic extrinsic calibration of a camera and a 3d lidar using line
  and plane correspondences.
\newblock In {\em IEEE/RSJ Intl. Conf. on Intelligent Robots and Systems,
  IROS}, October 2018.

\bibitem{Zhou2015a}
X.~Zhou, S.~Leonardos, X.~Hu, and K.~Daniilidis.
\newblock 3d shape reconstruction from 2d landmarks: A convex formulation.
\newblock In {\em IEEE Conf. Computer Vision and Pattern Recognition (CVPR)},
  2015.

\bibitem{Zhou2015b}
X.~Zhou, M.~Zhu, and K.~Daniilidis.
\newblock Multi-image matching via fast alternating minimization.
\newblock In {\em IEEE Int'l Conf. Computer Vision (ICCV)}, 2015.

\end{thebibliography}
}

\end{document}